\documentclass[10pt,twocolumn,letterpaper]{article}

\usepackage[accsupp]{axessibility} % Improves PDF readability for those with disabilities.

\usepackage[pagenumbers]{iccv} % To force page numbers, e.g. for an arXiv version

\definecolor{iccvblue}{rgb}{0.21,0.49,0.74}
\usepackage[pagebackref,breaklinks,colorlinks,allcolors=iccvblue,bookmarksopen]{hyperref}

\usepackage{times}
\usepackage{latexsym}
\usepackage{graphicx}
\usepackage{xspace} 
\usepackage[utf8]{inputenc} % allow utf-8 input
\usepackage[T1]{fontenc}    % use 8-bit T1 fonts
\usepackage{url}            % simple URL typesetting
\usepackage{booktabs}       % professional-quality tables
\usepackage{amsfonts}       % blackboard math symbols
\usepackage{nicefrac}       % compact symbols for 1/2, etc.
\usepackage{microtype}      % microtypography
\usepackage{color}
\usepackage{footnote}

\usepackage{bm}
\usepackage{bbm}
\usepackage{multirow}
\usepackage{pifont}
\usepackage{makecell}
\usepackage{mathtools}
\usepackage{colortbl} % 
\usepackage{enumitem} % enumerate
\usepackage{amsmath} % \text 
\usepackage{amssymb} %\blacktriangleright
\usepackage{adjustbox} 
\usepackage{tabularx} % tabularx
\usepackage{wrapfig,lipsum,booktabs}
\usepackage{soul} %\hl
\usepackage{xcolor}

\usepackage{tcolorbox}
\tcbuselibrary{skins}
\usepackage{tikz}
\usetikzlibrary{patterns}

\usetikzlibrary{arrows}
\usepackage{tkz-kiviat}
\usepackage{xfp}

\usepackage{algorithm}
\usepackage{algorithmicx}
\usepackage[noend]{algpseudocode}
\renewcommand{\algorithmicrequire}{\textbf{Input:}}
\renewcommand{\algorithmicensure}{\textbf{Output:}}

\newcommand{\bv}{\boldsymbol{b}}

\newcommand{\hv}{\boldsymbol{h}}

\newcommand{\nv}{\boldsymbol{n}}

\newcommand{\vv}{\boldsymbol{v}}

\newcommand{\alphav}{\boldsymbol{\alpha}}

\newtcbox{\mybox}[1][red]{on line, 
arc=6pt, 
outer arc=6pt,
colback = #1!10!white, 
colframe = #1!50!black,
boxsep=0pt, 
left=6pt, 
right=6pt, 
top=6pt, 
bottom=6pt,
boxrule=0.5pt, 
bottomrule=0.5pt, 
toprule=0.5pt
}
\newcommand{\myboxedtext}[2][red]{%
    \mybox[#1]{\parbox{\dimexpr\linewidth-16pt}{#2}}%
}

\newtcbox{\boxwofill}[1][red]{on line, 
arc=0pt, 
outer arc=0pt,
colback = white, 
colframe = #1,
boxsep=0pt, 
left=1pt, 
right=1pt, 
top=2pt, 
bottom=2pt,
boxrule=0.5pt, 
bottomrule=0.5pt, 
toprule=0.5pt
}

\definecolor{OliveGreen}{HTML}{75A140} % 
\definecolor{DeepBlueViolet}{HTML}{2C3284} % 
\definecolor{DeepGreen}{RGB}{40,174,127} % 
\definecolor{TealBlue}{HTML}{1AAAC2} % 
\definecolor{PurplishMauve}{HTML}{975AA4} % 
\definecolor{DarkBrown}{HTML}{715227} % 
\definecolor{LightCoralRed}{HTML}{D96161} % 
\definecolor{BrightOrange}{HTML}{F5AA42} % 
\definecolor{CoralOrange}{HTML}{F47750} % 
\definecolor{BrightGoldenYellow}{HTML}{F7CD3B} % 

\definecolor{GrayishBlue}{RGB}{108,116,175}
\definecolor{LightSkyBlue}{RGB}{123,207,241}
\definecolor{LightPeriwinkleBlue}{RGB}{152,177,241}
\definecolor{LightPurple}{RGB}{161,134,189}
\definecolor{LightMintGreen}{RGB}{152,212,171}
\definecolor{CoralPink}{RGB}{234,98,122}
\definecolor{LightCoralPink}{RGB}{251,210,206}
\definecolor{LightOrangeYellow}{RGB}{249,180,116}

\definecolor{DarkGray}{gray}{0.5}  %127,127,127
\definecolor{LighterGray}{gray}{0.9}
\definecolor{neutral_gray}{HTML}{D0CDCD}

\definecolor{dark_red}{HTML}{A10035}
\definecolor{back_red}{RGB}{255,190,190}
\definecolor{pale_red}{rgb}{0.90,0.61,0.58}  % 淡红色
\definecolor{neg_red}{HTML}{FFB0BA}  % 浅粉红色
\definecolor{brilliantrose}{rgb}{1.0,0.33,0.64} %玫瑰红色
\definecolor{carminepink}{rgb}{0.92, 0.3, 0.26}  % 胭脂红
\definecolor{red_bright}{HTML}{FF0000}   % Bright Red
\definecolor{red_wine}{HTML}{A10035}     % Wine Red
\definecolor{red_crimson}{HTML}{DC143C}  % Crimson
\definecolor{red_firebrick}{HTML}{B22222}% Firebrick
\definecolor{red_dark}{HTML}{8B0000}     % Dark Red
\definecolor{red_indian}{HTML}{CD5C5C}   % Indian Red
\definecolor{red_tomato}{HTML}{FF6347}   % Tomato
\definecolor{red_orange}{HTML}{FF4500}   % Orange Red
\definecolor{red_salmon}{HTML}{FA8072}   % Salmon
\definecolor{red_darksalmon}{HTML}{E9967A}% Dark Salmon

\definecolor{pale_green}{rgb}{0.55,0.75,0.60} % 淡绿色
\definecolor{mygreen}{HTML}{009B55}  % green
\definecolor{ggreen}{rgb}{0.0,0.5,0.0}
\definecolor{forestgreen}{rgb}{0.13, 0.55, 0.13}
\definecolor{darkgreen}{HTML}{448E64}
\definecolor{pos_green}{HTML}{CAECD0}
\definecolor{green_bright}{HTML}{00FF00}    % Bright Green
\definecolor{green_forest}{HTML}{228B22}    % Forest Green
\definecolor{green_dark}{HTML}{006400}      % Dark Green
\definecolor{green_lime}{HTML}{32CD32}      % Lime Green
\definecolor{green_olive}{HTML}{808000}     % Olive
\definecolor{green_sea}{HTML}{2E8B57}       % Sea Green
\definecolor{green_mediumseagreen}{HTML}{3CB371} % Medium Sea Green
\definecolor{green_pale}{HTML}{98FB98}      % Pale Green
\definecolor{green_spring}{HTML}{00FF7F}    % Spring Green
\definecolor{green_mint}{HTML}{98FF98}      % Mint Green
\definecolor{green_jungle}{HTML}{29AB87}    % Jungle Green
\definecolor{green_teal}{HTML}{008080}      % Teal

\definecolor{aliceblue}{rgb}{0.94,0.97,1.0}
\definecolor{backblue}{RGB}{210,230,250}
\definecolor{NiceBlue}{rgb}{0.11764705882352941, 0.5647058823529412, 1.0}
\definecolor{gray_blue}{rgb}{0.52,0.59,0.69}  %云灰蓝
\definecolor{azure}{rgb}{0.0, 0.5, 1.0}  %天蓝色 
\definecolor{blue_light_sky}{RGB}{76,137,237}
\definecolor{blue_ele_sky}{RGB}{57,122,228}

\definecolor{blue_bright}{HTML}{0000FF}     % Bright Blue
\definecolor{blue_sky}{HTML}{87CEEB}        % Sky Blue
\definecolor{blue_deepsky}{HTML}{00BFFF}    % Deep Sky Blue
\definecolor{blue_royal}{HTML}{4169E1}      % Royal Blue
\definecolor{blue_midnight}{HTML}{191970}   % Midnight Blue
\definecolor{blue_dodger}{HTML}{1E90FF}     % Dodger Blue
\definecolor{blue_steel}{HTML}{4682B4}      % Steel Blue
\definecolor{blue_cadet}{HTML}{5F9EA0}      % Cadet Blue
\definecolor{blue_cornflower}{HTML}{6495ED} % Cornflower Blue
\definecolor{blue_light}{HTML}{ADD8E6}      % Light Blue
\definecolor{blue_powder}{HTML}{B0E0E6}     % Powder Blue
\definecolor{blue_navy}{HTML}{000080}       % Navy Blue

\definecolor{LightCyan}{HTML}{bbe2fc} % {rgb}{0.98,0.88,0.82}
\definecolor{NiceCyan}{HTML}{3db0fc} 
\definecolor{NiceIIcyan}{RGB}{112,207,224}
\definecolor{cyan_steel}{RGB}{14,111,199}
\definecolor{dark_pink}{RGB}{112, 176, 215}

\definecolor{cyan_bright}{HTML}{00FFFF}    % Bright Cyan
\definecolor{cyan_light}{HTML}{E0FFFF}     % Light Cyan
\definecolor{cyan_dark}{HTML}{008B8B}      % Dark Cyan
\definecolor{cyan_aqua}{HTML}{7FFFD4}      % Aqua Cyan (Aquamarine)
\definecolor{cyan_teal}{HTML}{008080}      % Teal Cyan
\definecolor{cyan_turquoise}{HTML}{40E0D0} % Turquoise Cyan
\definecolor{cyan_medium}{HTML}{48D1CC}    % Medium Cyan (Medium Turquoise)
\definecolor{cyan_pale}{HTML}{AFEEEE}      % Pale Cyan (Pale Turquoise)

\definecolor{challenge_yellow}{HTML}{FFD451}
\definecolor{gold_bright}{HTML}{FFD700}   % Bright Gold
\definecolor{gold_dark}{HTML}{B8860B}     % Dark Gold (Dark Goldenrod)
\definecolor{gold_light}{HTML}{FFFACD}    % Light Gold (Lemon Chiffon)
\definecolor{gold_champagne}{HTML}{F7E7CE} % Champagne Gold
\definecolor{gold_amber}{HTML}{FFBF00}    % Amber Gold
\definecolor{gold_sunset}{HTML}{FFCC33}   % Sunset Gold
\definecolor{gold_pale}{HTML}{EEE8AA}     % Pale Gold (Pale Goldenrod)
\definecolor{gold_royal}{HTML}{F4C430}    % Royal Gold

\definecolor{violet}{HTML}{BF00FF} %CC00CC,BF00FF
\definecolor{brightlavender}{rgb}{0.75, 0.58, 0.89}  %薰衣草
\definecolor{purple_bright}{HTML}{800080}   % Bright Purple
\definecolor{purple_lavender}{HTML}{E6E6FA} % Lavender Purple
\definecolor{purple_dark}{HTML}{4B0082}     % Dark Purple (Indigo)
\definecolor{purple_violet}{HTML}{9400D3}   % Violet Purple (Dark Violet)
\definecolor{purple_thistle}{HTML}{D8BFD8}  % Thistle Purple
\definecolor{purple_plum}{HTML}{DDA0DD}     % Plum Purple
\definecolor{purple_magenta}{HTML}{FF00FF}  % Magenta Purple (Fuchsia)
\definecolor{purple_heather}{HTML}{9966CC}  % Heather Purple (Amethyst)
\definecolor{electric_purple}{RGB}{146,36,171}

\definecolor{bk-tcolorbox}{RGB}{242,242,242}

\definecolor{dark_emerald}{RGB}{28,127,60} 
\definecolor{emerald}{HTML}{50C878}     % Emerald Green
\definecolor{olive_green}{HTML}{808000} % Olive Green
\definecolor{coral}{HTML}{FF7F50}       % Coral
\definecolor{blush_pink}{HTML}{FFC0CB}  % Blush Pink
\definecolor{indigo}{HTML}{4B0082}      % Indigo
\definecolor{turquoise}{HTML}{40E0D0}   % Turquoise
\definecolor{slate_gray}{HTML}{708090}  % Slate Gray
\definecolor{dark_orange}{HTML}{FF8C00} % Dark Orange
\definecolor{rose_gold}{HTML}{B76E79}   % Rose Gold
\definecolor{peacock_blue}{HTML}{1C39BB} % Peacock Blue
\definecolor{magenta}{HTML}{FF00FF}     % Magenta
\definecolor{royal_blue}{HTML}{4169E1}  % Royal Blue
\definecolor{cranberry}{HTML}{9E003A}   % Cranberry Red

\DeclareRobustCommand{\hlp}[1]{{\sethlcolor{pos_green}\hl{#1}}}

\definecolor{highlight_text}{RGB}{255,255,255}
\definecolor{smk_reason}{RGB}{73,168,100}

\definecolor{tfc_understand}{RGB}{156,45,177}

\definecolor{ocr_understand}{RGB}{245,182,53}

\definecolor{text_only}{RGB}{76,137,237}

\newcommand{\listnumber}[1]{{#1}}

\newcommand{\pgraph}[1]{\noindent \textbf{#1.\,}}

\definecolor{color_our}{HTML}{e6f2c2}
\definecolor{NiceGray}{HTML}{696969}
\definecolor{NiceGreen}{HTML}{b6c783}

\usepackage[capitalize]{cleveref}
\crefname{section}{Sec.}{Secs.}
\Crefname{section}{Section}{Sections}
\Crefname{table}{Table}{Tables}
\crefname{table}{Table}{Tables}
\Crefname{figure}{Figure}{Figures}
\crefname{figure}{Figure}{Figures}

\Crefname{theorem}{Theorem}{Theorems}
\crefname{theorem}{Thm.}{Thm.s}
\Crefname{algorithm}{Algorithm}{Algorithms}
\crefname{algorithm}{Algo.}{Algos.}
\crefname{appendix}{App.}{Apps.}    
\Crefname{appendix}{Appendix}{Appendices}

\newcommand{\Ours}{Corvid\xspace}
\newcommand{\mmconnector}{GateMixer\xspace}
\newcommand{\mmcot}{{MCoT-Instruct}\xspace}
\definecolor{darkgreen}{rgb}{0.0, 0.8, 0.0}

\def\paperID{11749} % * Enter the Paper ID here
\def\confName{ICCV}
\def\confYear{2025}

\title{{\Ours}\includegraphics[height=15pt]{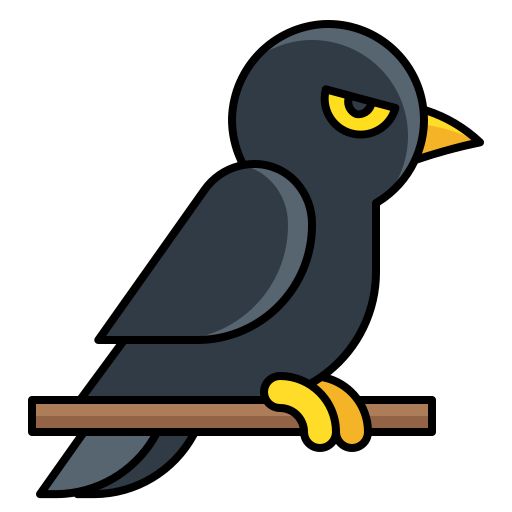}: Improving Multimodal Large Language Models Towards\\ Chain-of-Thought Reasoning}

\author{Jingjing Jiang$^{1,2}$,~Chao Ma$^{1}$,~Xurui Song$^{2}$,~Hanwang Zhang$^{2}$,~Jun Luo$^{2}$
\vspace{0.2cm}
\\
$^{1}$Shanghai Jiao Tong University, $^{2}$Nanyang Technological University
\vspace{0.2cm}
\\
}

\begin{document}
\maketitle

\begin{abstract}
Recent advancements in multimodal large language models (MLLMs) have demonstrated exceptional performance in multimodal perception and understanding. However, leading open-source MLLMs exhibit significant limitations in complex and structured reasoning, particularly in tasks requiring deep reasoning for decision-making and problem-solving. In this work, we present Corvid, an MLLM with enhanced chain-of-thought (CoT) reasoning capabilities. Architecturally, Corvid incorporates a hybrid vision encoder for informative visual representation and a meticulously designed connector (GateMixer) to facilitate cross-modal alignment. To enhance Corvid's CoT reasoning capabilities, we introduce MCoT-Instruct-287K, a high-quality multimodal CoT instruction-following dataset, refined and standardized from diverse public reasoning sources. Leveraging this dataset, we fine-tune Corvid with a two-stage CoT-formatted training approach to progressively enhance its step-by-step reasoning abilities. Furthermore, we propose an effective inference-time scaling strategy that enables Corvid to mitigate over-reasoning and under-reasoning through self-verification. Extensive experiments demonstrate that Corvid outperforms existing o1-like MLLMs and state-of-the-art MLLMs with similar parameter scales, with notable strengths in mathematical reasoning and science problem-solving. 
Project page: \url{https://mm-vl.github.io/corvid}.
\end{abstract}

%%%%%%%%%%%%%%%%%%%%%%%%%%%%%%%%%%%%%%%%%%%%%%%%%%%%%%%%%%%%
%%%%%%%%% BODY TEXT
\section{Introduction}
\label{sec:intro}

% ************************************************************************
\begin{figure}[!t]
\centering
\includegraphics[width=0.99\linewidth]{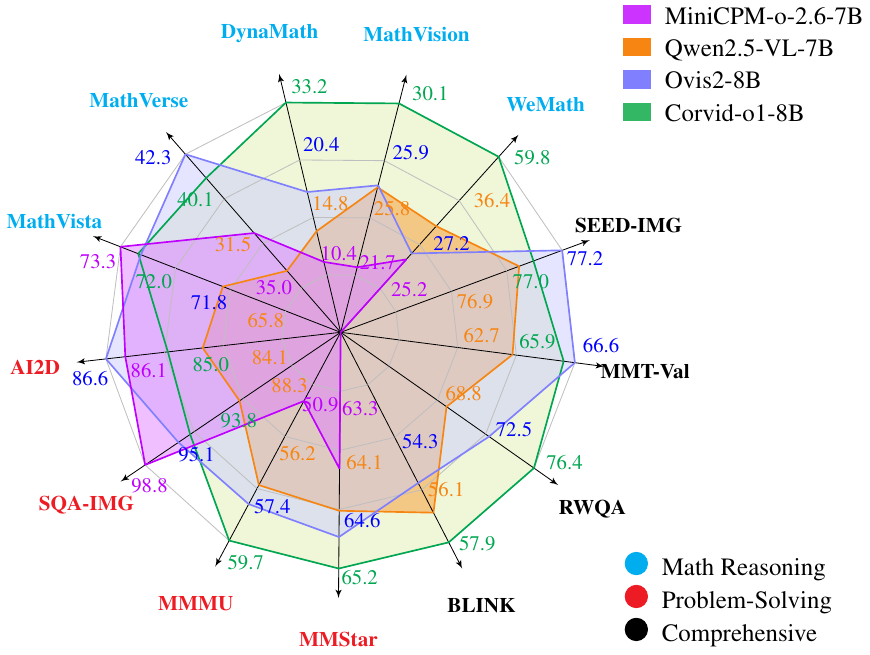}
%
% \raggedleft %Important!!!
% \resizebox{1.0\linewidth}{!}{
% \hspace{-88pt}
% \raggedleft %Important!!!
% \input{figure/radar_corvid.tex}
% }
%
\vspace{-2mm}
\caption{
\textbf{Comparison with leading open-source MLLMs} with on-par parameter scales. \Ours-o1-8B showcases superiority in mathematical reasoning and problem-solving. 
}
\label{fig:infer_exp}
\vspace{-2mm}
\end{figure}
\begin{table}[t]
\footnotesize
\centering
\setlength{\tabcolsep}{.55mm}{
\begin{tabularx}{\linewidth}{lccccccc}
\toprule 
{MLLMs} 
&{MMStar} 
&{MMB} 
&{MMVet} 
&{MathV} 
&{AI2D} 
&{Hallusion} 
&{Avg.} 
\\ 
\midrule
% Llama-3.2-11B-Vision-Instruct (baseline)~\cite{llama32v}
Baseline
&49.8 &65.8 &57.6 &48.6 &77.3 &40.3 &56.9
\\ 
\quad LLaVA-o1
% &57.6 &75.0 &60.3 &54.8 &85.7 &47.8 &63.5
&58.1 &75.6 &61.7 &56.1 &78.8 &48.2 &63.1 % citing from paper 
\\ 
\quad LlamaV-o1
&59.5 & 79.9 & 65.4 & 54.4 &81.2 &\textbf{63.5} & 67.3
\\
\rowcolor{ggreen!20}
\quad \textbf{\Ours-o1\textsuperscript{\textdagger}}
&62.2 &80.3 &\textbf{67.0} &61.5 &81.2 &58.7 &\textbf{68.5}
\\ 
\midrule
Mulberry-o1-7B
&61.3 &75.3 &43.9 &57.5 &79.0 &54.1 &62.8
\\ 
\rowcolor{ggreen!20}
\textbf{\Ours-o1-8B} 
&\textbf{65.2} &\textbf{82.9} &45.9 &\textbf{72.0} &\textbf{85.0} &54.0 &\textbf{67.5} 
\\ 
\bottomrule
\end{tabularx}
}
\vspace{-2mm}
\caption{\textbf{Compared with o1-Like MLLMs}, \Ours demonstrates superior overall performance across multiple benchmarks. 
Here, \Ours-o1\textsuperscript{\textdagger}, LLaVA-o1, and LlamaV-o1 utilize the same baseline MLLM, Llama-3.2-11B-Vision-Instruct. 
}
\label{tab:bench}
\vspace{-2mm}
\end{table}

% ************************************************************************

Multimodal large language models (MLLMs)~\cite{liu2024llavanext,lu2024ovis,bai2025qwen2.5-vl,chen2024expanding,yao2024minicpm,zhang2024mm1} have demonstrated exceptional capabilities in perception and understanding by integrating visual modalities into powerful large language models (LLMs)~\cite{team2024gemma,touvron2023llama,bai2023qwenllm}. 
However, as illustrated in \cref{fig:infer_exp}, leading MLLMs, such as Ovis2~\cite{lu2024ovis} and Qwen2.5-VL~\cite{bai2025qwen2.5-vl}, still exhibit suboptimal performance on complex tasks requiring \textit{deep thinking and extrapolation} for effective problem-solving. 
Pioneering o1-like MLLMs~\cite{luo2025ursa,xu2024llava,thawakar2025llamav,yao2024mulberry} solve such sophisticated tasks through enhancing chain-of-thought (CoT) reasoning~\cite{wei2022chain} capability, which explicitly unveils logical thought processes and generates step-by-step rationales before arriving at outcomes. 
Despite these advancements, MLLMs still encounter challenges in complex and structured reasoning.

In this work, we aim to further enhance MLLMs for CoT reasoning by addressing three critical challenges. 
\textit{\textbf{First}, there remains a significant shortage of high-quality multimodal CoT data.} Recent empirical studies~\cite{zhang2024improve} have demonstrated that MLLMs trained using direct responses cannot perform step-by-step reasoning, highlighting the urgent need for multimodal CoT data. However, manually created CoTs are typically brief, while AI-generated examples tend to be noisy, rendering them unsuitable for effective CoT-formatted training. 
\textit{\textbf{Second}, MLLMs frequently reason using flawed visual evidence due to insufficient representation and misalignment.} To achieve complex cross-modal reasoning, MLLMs must accurately capture visual information and efficiently transform it into language embedding space, providing the condition and rationale for LLM reasoning. Therefore, optimizing visual representation and cross-modal alignment in MLLMs is essential for further enhancing their reasoning capabilities. 
\textit{\textbf{Third}, MLLMs are prone to over-reasoning and under-reasoning during inference.} Current o1-like MLLMs execute deep reasoning uniformly across all testing instances, regardless of task complexity. In practice, complex reasoning is not always necessary for correct outcomes, particularly for straightforward tasks where inference directly may yield more accurate answers than CoT reasoning. This limitation primarily stems from the context loss and hallucination of MLLMs during long-chain generation.

To address these challenges, we develop an MLLM with advanced CoT reasoning capabilities, referred to as \textbf{Corvid}. 
Architecturally, Corvid integrates two visual foundation models as a hybrid vision encoder for sufficient visual representation and incorporates a novel connector, \textbf{GateMixer}, to facilitate modality interaction and alignment through a gate mechanism with selective attention. 
To enhance Corvid's CoT reasoning capabilities, we first refine and standardize multiple multimodal reasoning datasets spanning diverse reasoning types and domains, resulting in a high-quality multimodal CoT instruction-following dataset, \textbf{MCoT-Instruct-287K}. Building upon this foundation, we meticulously curate three specialized datasets to support Corvid's training and optimization. 
Subsequently, Corvid undergoes a two-stage CoT-formatted training process following its alignment pre-training, to progressively develop step-by-step reasoning while ensuring comprehensive visual understanding. 
Finally, we propose scaling inference-time computation to mitigate the issues of over-reasoning and under-reasoning. Specifically, we introduce an effective \textbf{self-verification} strategy that enables the model to determine whether to perform CoT reasoning conditioned on task complexity.

Corvid is implemented based on the open-source Llama3-8B-Instruct~\cite{dubey2024llama} and evaluated across various benchmarks involving mathematical reasoning, science problem-solving, and comprehensive capabilities. As illustrated in \cref{fig:infer_exp} and \cref{tab:bench}, Corvid outperforms leading open-source MLLMs with on-par parameter scales on 7 out of 13 benchmarks and showcases substantial performance advantages over existing o1-like MLLMs across multiple benchmarks. Furthermore, we conduct systematic ablation studies and analyses to validate the effectiveness and contributions of individual Corvid components. 
Our main contributions are as follows: 
\begin{itemize}
\item We develop Corvid, an MLLM with advanced CoT reasoning capabilities. 
It incorporates a hybrid vision encoder for sufficient visual representation and a novel connector (GateMixer) for enhanced cross-modal alignment. 
\item We propose an effective inference-time scaling strategy, enabling MLLMs to alleviate over-reasoning and under-reasoning through self-verification. 
\item We introduce MCoT-Instruct-287K, a high-quality multimodal CoT dataset covering diverse reasoning types. 
\item Extensive experiments demonstrate the superiority of Corvid against leading open-source MLLMs of comparable parameter sizes and existing o1-like MLLMs. 
\end{itemize}

% ************************************************************************** 
\begin{figure*}[ht!]
\centering 
\includegraphics[width=0.9\textwidth]{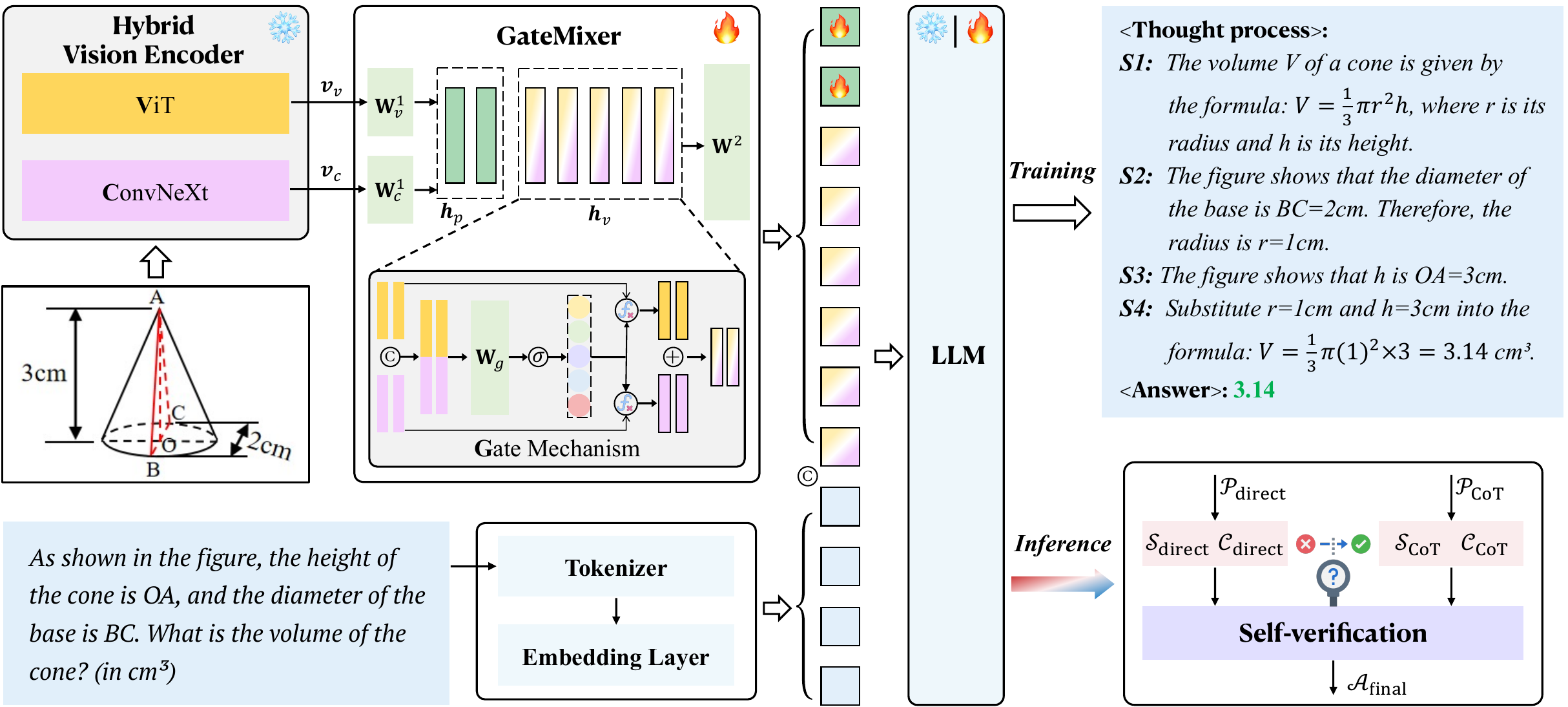}
\vspace{-2mm}
\caption{\textbf{Model Overview.} 
\Ours incorporates a hybrid vision encoder to effectively represent visual content and a specially designed connector (\mmconnector) to enhance alignment with the LLM. 
During inference, \Ours implements a self-verification strategy to mitigate over-reasoning and under-reasoning, ensuring more balanced and accurate responses. 
}
\label{fig:lmm_overview}
\vspace{-2mm}
\end{figure*}
% ************************************************************************** 

\section{Related Work}
\label{sec:related}

\pgraph{Modality Alignment in MLLMs}
MLLMs~\cite{chen2024internvl,lu2024ovis,bai2025qwen2.5-vl,yao2024minicpm,liu2024llavanext} aim to extend the exceptional capabilities of LLMs to complex tasks involving multiple modalities beyond text. 
%To achieve this, MLLMs first require the architectural integration of visual foundation models with LLMs through specialized modules (\ie, \textit{connectors}) to establish cross-modal semantic links. 
To achieve this, MLLMs require specialized modules (\ie, \textit{connectors}) to integrate visual foundation models with LLMs. 
The most common MLP-based connector~\cite{tsimpoukelli2021multimodal,driess2023palm,zhao2023mllm,gao2023llamaadapterv2,liu2023visual,luo2023cheap} utilizes simple linear projection layers to map visual features into the textual embedding space, which is lightweight but short of interaction. In contrast, another line of work~\cite{li2023blip2,dai2023instructblip,alayrac2022flamingo,gong2023multimodal,ye2023mplug} incorporates cross-attention layers into the connector to facilitate interaction between modalities. 
Once structural connections are established, the next objective is to enable LLMs to understand visual concepts by learning from cross-modal calibrated data. A feasible practice is \textit{alignment pre-training}, which involves training the connector on large-scale image-text pairs, such as ShareGPT4V~\cite{chen2023sharegpt4v} and PixMo~\cite{deitke2024molmo}. 
Similarly, our GateMixer, with gate attention, is trained on multi-grained calibrated data with enhanced cross-modal alignment.

\vspace{2pt}
\pgraph{CoT Reasoning with MLLMs} 
Chain-of-thought refers to a series of intermediate reasoning steps or rationales for deriving the final outcome~\cite{wei2022chain} and has been extensively demonstrated to elicit the powerful reasoning capabilities of LLMs~\cite{cheng2024chainlm,fu2023chain,wang2023self,diao2023active}. 
Multimodal CoT reasoning extends this concept by leveraging \textit{CoT prompting}~\cite{gao2024cantor,mitra2023compositional,lu2023chameleon} and \textit{CoT-formatted tuning}~\cite{luo2025ursa,xu2024llava,thawakar2025llamav} to perform more complex multimodal tasks, such as mathematical reasoning~\cite{zhang2024mavis} and robot planning~\cite{mu2023embodiedgpt}. 
CoT prompting is typically applied under zero-shot~\cite{kojima2022large} or few-shot~\cite{zhang2023automatic} paradigms, where large multimodal models like GPT-4o~\cite{openai2024gpt4o} and Claude-3.5-Sonnet~\cite{anthropic2024claude} engage in step-by-step thinking before reaching an outcome. 
On the other hand, CoT-formatted tuning is the \textit{visual instruction tuning} of MLLMs using multimodal CoT instruction-following datasets, where performance hinges on the quality of CoT responses. 
Existing multimodal CoT data primarily originate from manual creation~\cite{zellers2019recognition,lu2022learn,schwenk2022okvqa} and AI-assisted generation~\cite{zhao2023mllm,xu2024llava,zhang2024mavis}. 
Manually-created CoTs are usually accurate but brief, while generated CoTs tend to be more detailed but may contain errors and duplications. 
In this work, we separately refine and standardize manually-created and AI-generated reasoning datasets with GPT assistance to obtain high-quality CoT data.

\vspace{2pt}
\pgraph{Inference-time Scaling in MLLMs}
Inference-time scaling~\cite{snell2024scaling} seeks to further enhance model performance by enabling iterative reasoning, allowing models to generate multiple responses, engage in self-reflection, and refine their answers through exploratory problem-solving strategies. Recent studies, such as DeepSeek-R1~\cite{deepseek2025deepseek} and OpenAI-o1~\cite{jaech2024openai}, have demonstrated the effectiveness of inference-time scaling techniques~\cite{havrilla2024glore,wang2022self,huang2022large,weng2022large} for depth reasoning in LLM. However, the application of inference-time scaling for MLLM reasoning remains relatively unexplored. 
Recent approaches have attempted to exploit this technique to improve multimodal reasoning. For instance, LLaVA-o1~\cite{xu2024llava} and LlamaV-o1~\cite{thawakar2025llamav} utilize stage-level and sentence-level beam search, respectively, to generate multiple reasoning paths and select the optimal one. Similarly, Mulberry-o1~\cite{yao2024mulberry} introduces a collective Monte Carlo tree search algorithm for reasoning path selection. 
In contrast, we propose an inference-time self-verification strategy to select the final answer without relying on any external verifier.

\section{Methodology}  
\label{sec:method}

\subsection{Model Architecture}  
\label{sec:model_architecture}

As illustrated in \cref{fig:lmm_overview}, \Ours comprises the following three crucial modules:

\vspace{2pt}
\pgraph{Hybrid Vision Encoder} 
Visual understanding is fundamental for MLLMs to perform cross-modal reasoning. To maximize this capability in \Ours, we strategically integrate the pretrained SigLIP ViT-SO400M~\cite{zhai2023sigmoid} and OpenCLIP ConvNeXt-XXL~\cite{liu2022convnet} as a hybrid vision encoder, guided by recent investigations on optimal vision encoders for MLLMs~\cite{shi2024eagle,wei2024vary,tong2024cambrian}. These studies empirically demonstrate that SigLIP and ConvNeXt-XXL represent the state-of-the-art ViT-based and CNN-based encoders across various multimodal benchmarks. 
Specifically, the ViT encoder processes input images at a resolution of $384 \times 384$ to extract semantically rich features $\vv_{\emph{v}} \in \mathbb{R}^{729 \times 1152}$, while the ConvNeXt encoder operates the same image at the same resolution to generate multi-stage aggregated features $\vv_\emph{c} \in \mathbb{R}^{729\times 5760}$ that preserve spatial details.

\vspace{2pt}
\pgraph{LLM} 
\Ours primarily employs the open-source Llama3-8B~\cite{dubey2024llama} as its LLM decoder, which encodes concatenated embeddings of visual and language tokens to generate instruction-following CoT responses.

% ************************************************************************** 
\begin{table*}[!th]
\centering
\footnotesize 
\renewcommand{\arraystretch}{1.5}
\setlength{\tabcolsep}{1.2mm}{
\begin{tabular}{@{}c|>{\raggedleft\arraybackslash}m{3.4cm}|>{\centering\arraybackslash}p{11.8cm}}
\toprule
Name 
&\multicolumn{1}{c|}{Data Type (Proportion)}
&\multicolumn{1}{c}{Source Datasets}
\\ 
\midrule
\rowcolor{gray!20} \cellcolor{white!20}
&Coarse-grained (32.1\%) 
&LLaVA-Pretrain~\cite{liu2024improvedllava} (321K)
\\
&\multirow{2}{*}{Fine-grained (32.1\%)} 
&ALLaVA-4V~\cite{chen2024allava} (195K), Docci~\cite{onoe2024docci} (15K), ShareGPT-4o~\cite{cui2024sharegpt4o} (49K), ShareGPT4V~\cite{chen2023sharegpt4v} (52K), VG~\cite{krishna2017visual}~(10K)
\\
\rowcolor{gray!20} \cellcolor{white!20}
\multirow{-4}{*}{MGA-1M}
&Chart, Math, OCR (35.8\%)
&ChartCap~\cite{kantharaj2022chart} (30K), MAVIS-Cap~\cite{zhang2024mavis} (306K), TextCaps~\cite{sidorov2020textcaps} (22K)
\\

\midrule
&CoT Reasoning (20.2\%)
&\textbf{MCoT-Instruct} (124K), MAVIS-Instruct~\cite{zhang2024mavis} (81K) 
\\
\rowcolor{PurplishMauve!20} \cellcolor{white!20}
&\multirow{2}{*}{Direct Reasoning (31.3\%)}
&AI2D~\cite{kembhavi2016diagram} (4K), CLEVR~\cite{johnson2017clevr} (70K), CLEVR-Math~\cite{lindstrom2022clevr} (85K), HatefulMemes~\cite{kiela2020hateful} (8K), VSR~\cite{liu2022visual} (2K), IconQA~\cite{lu2021iconqa} (27K), Inter-GPS~\cite{lu2021inter} (1K), RAVEN~\cite{zhang2019raven} (21K), TallyQA~\cite{acharya2019tallyqa} (99K), TQA~\cite{alawwad2024enhancing} (1K)
\\ 

&\multirow{2}{*}{Chart, Figure, Table (29.7\%)}
&ChartQA~\cite{masry2022chartqa} (18K), DVQA~\cite{kafle2018dvqa} (49K), FigureQA~\cite{kahou2017figureqa} (49K), HiTab~\cite{cheng2021hitab} (2K), MapQA~\cite{chang2022mapqa} (37K), PlotQA~\cite{methani2020plotqa} (49K), SQA~\cite{iyyer2017search} (9K), TAT-QA~\cite{zhu2021tat} (2K), WikiSQL~\cite{zhong2017seq2sql} (49K), WTQ~\cite{pasupat2015compositional} (38K)
\\
\rowcolor{PurplishMauve!20} \cellcolor{white!20}
&\multirow{2}{*}{OCR, Doc (13.8\%)} 
&IAM~\cite{marti2002iam} (6K), OCRVQA~\cite{mishra2019ocr} (80K), InfoGraphicVQA~\cite{mathew2022infographicvqa} (2K), TextVQA~\cite{singh2019towards} (22K), Visualmrc~\cite{tanaka2021visualmrc} (3K), DocVQA~\cite{mathew2021docvqa} (10K), ST-VQA~\cite{biten2019scene} (17K)
\\  
\multirow{-8}{*}{\Ours-1M}
&Language-only (5.0\%)
&CamelMath~\cite{li2023camel} (12K), Dolly~\cite{mike2023free} (15K), Orca-Math~\cite{mitra2024orca} (13K), OpenHermes-2.5~\cite{openhermes} (12K)
\\

\midrule
\rowcolor{teal!20} \cellcolor{white!20}
o1-320K
&CoT Reasoning (100\%)
&\textbf{MCoT-Instruct} (163K), MAVIS-Instruct~\cite{zhang2024mavis} (137K), CamelMath~\cite{li2023camel} (20K) 
\\
\bottomrule
\end{tabular}
}
\vspace{-2mm}
\caption{\textbf{Summary of training data.} 
\textbf{MCoT-Instruct} is our high-quality multimodal CoT instruction-following dataset, refined and standardized from diverse publicly available reasoning datasets (see Supplementary Material for complete details). 
Numbers in parentheses indicate the number of conversation instances used for each dataset. 
}
\label{tab:training_data}
\vspace{-2mm}
\end{table*}

% ************************************************************************** 

\vspace{2pt}
\pgraph{\mmconnector} 
To better align the hybrid vision encoder with LLMs, we design a novel connector to transform $\vv_{\emph{v}}$ and $\vv_{\emph{c}}$ into LLM-friendly token embeddings. 
As shown in \cref{fig:lmm_overview}, the connector initially employs two distinct linear layers $\mathbf{W}^1_\emph{v}$ and $\mathbf{W}^1_\emph{c}$ to map $\{\vv_\emph{v}, \vv_\emph{c}\}$ into a unified embedding space, resulting in $\{\hv_\emph{v}, \hv_\emph{c}\} \in \mathbb{R}^{729\times d}$. 
To maintain the current length of visual tokens without significant alteration, we take insight from the input gate mechanism in LSTM~\cite{hochreiter1997long}, and thus propose to element-wisely mix $\hv_\emph{v}$ and $\hv_\emph{c}$ using a gate mechanism with selective attention, \ie, 
\begin{align}
&\alphav = \sigma(\mathbf{W}_\emph{g}[\hv_\emph{v}; \,\hv_\emph{c}] + \bv_\emph{g}), \\ 
&\hv = (1 - \alphav) \odot \hv_\emph{v} + \alphav \odot \hv_\emph{c}, 
\label{eq:gate_attn}
\end{align}
where $\mathbf{W}_\emph{g} \in \mathbb{R}^{d \times 2d}$, $\bv_\emph{g} \in \mathbb{R}^{d}$, $\sigma$ denotes the Sigmoid function, $\odot$ and $[ ;]$ respectively represent the operations of element-wise matrix multiplication and vector concatenation. 
Subsequently, we sequence-wisely insert learnable prefix token embeddings $\hv_{\emph{p}} \in \mathbb{R}^{N_\emph{p} \times d}$ at the beginning of $\hv$ to facilitate context capture and generalization. 
Finally, the integrated visual features are projected into the language embedding space via a linear projection layer $\mathbf{W}^2$, resulting in $\hv_\emph{img}^0 \in \mathbb{R}^{(N_\emph{p} + 729) \times d}$. With the proposed \mmconnector, \Ours can enhance visual representations for correlated vision-language alignment.

% ************************************************************************** 
\subsection{Training Data Curation} 
\label{sec:data_curation}

Recent studies~\cite{tong2024cambrian,mckinzie2024mm1} have demonstrated the pivotal role of data quality and scale in improving MLLM performance. As a result, we introduce three meticulously curated datasets to support the training and optimization of MLLMs.

% ************************************************************************** 
\vspace{2pt}
\pgraph{MAG-1M} 
Beyond the alignment module, cross-modal calibrated data is essential for MLLMs to establish cross-modal semantic links. 
To achieve multi-grained alignment between vision and language components, we select nine publicly available image-text pair datasets, as listed in \cref{tab:training_data}, for alignment pretraining. 
Here, coarse-grained data supports image-level alignment, fine-grained data facilitates object-level alignment, and the chart, math, and OCR datasets contribute to data diversity.  

% ************************************************************************** 
\vspace{2pt}
\pgraph{\Ours-1M}
We curate a 1M instruction-following dataset encompassing a diverse range of data types to facilitate supervised fine-tuning. As summarized in \cref{tab:training_data}, this dataset includes approximately 51.5\% reasoning data, with CoT-formatted reasoning data comprising 20.2\%, ensuring that MLLMs effectively learn to reason. To enable accurate extraction of visual evidence for reasoning, we augment the dataset using specialized data with various image types, including charts, figures, tables, OCR, and documents, as MLLMs frequently extract critical information from such images for complex reasoning in science, mathematics, and logic. Finally, we incorporate a small portion of language-only instruction-following data to mitigate catastrophic forgetting during model training. 

% ************************************************************************** 
\vspace{2pt}
\pgraph{o1-320K} 
To further strengthen the o1-like reasoning capabilities of MLLMs, we curate a high-quality instruction dataset in a pure CoT format. As shown in \cref{tab:training_data}, this dataset comprises two multimodal CoT datasets (\mmcot, MAVIS-Instruct) and one language-only CoT dataset, totaling 320K samples. Here, \mmcot is our newly refined CoT dataset, featuring standardized and detailed reasoning steps as well as accurate final reasoning outcomes. 
%  (see \cref{app:mcot_instruct} for more details).

% ************************************************************************** 
\subsection{Training Procedure}
\label{sec:training_procedure}

We train our model in the following three consecutive stages:

\vspace{2pt}
\pgraph{Stage 1: Multi-Grained Alignment Pre-training} 
In this stage, we freeze all parameters of the hybrid vision encoder and LLM, training only the \mmconnector on MGA-1M to enable \Ours to establish conceptual links between visual and linguistic elements within the embedding space. In addition to the text generation loss, we introduce a contrastive regularization $\mathcal{L}_\text{CReg}$ to further promote semantic correlation between image and text representations, \ie, 
\begin{align}
\mathcal{L}_\text{CReg} = -\frac{1}{2b}\sum_{i=1}^{b}\left[\log\frac{\mathcal{S}_{ii}}{\sum_{j=1}^{b} \mathcal{S}_{ji}} + \log\frac{\mathcal{S}_{ii}}{\sum_{j=1}^{b} \mathcal{S}_{ij}} \right],
\label{eq:loss_pt}
\end{align}
where $b$ denotes the batch size, $\mathcal{S}$ is the similarity measure between the average-pooled image and text representations (\ie, $\bar{\hv}_\emph{img} \in \mathbb{R}^{b \times d}$ and $\bar{\hv}_\emph{txt} \in \mathbb{R}^{b \times d}$) after LLM. Throughout this work, we employ cosine similarity as the measure. 
% Thus, the total training loss in this stage is $\mathcal{L}_\text{TGen} + \mathcal{L}_\text{CReg}$. 

% ************************************************************************** 
\vspace{2pt}
\pgraph{Stage 2: CoT-Enhanced Supervised Fine-tuning}
In the second stage, we jointly train the proposed \mmconnector and the LLM on \Ours-1M. Upon completion of this stage, we obtained \textbf{\Ours-base}, the foundational model capable of following instructions and performing CoT reasoning. 

% ************************************************************************** 
\vspace{2pt}
\pgraph{Stage 3: Pure-CoT Instruction Tuning}
After enabling \Ours with cross-modal perception abilities and basic CoT reasoning capabilities in the preceding two stages, this stage aims to further strengthen its capability for systematic and structured reasoning. 
To this end, we fine-tune \Ours-base on o1-320K, resulting in \textbf{\Ours-o1}, the final model tailored for complex multimodal reasoning.

% ************************************************************************** 
\subsection{Model Inference} 
\label{sec:inference}

To mitigate \textit{over-reasoning on easy samples} and \textit{under-reasoning on hard ones} during inference, we propose a \textbf{self-verification} strategy for handling tasks of varying difficulty. 
As summarized in Algorithm~\ref{alg:algorithm1}, this strategy first prompts the model to generate both a direct response, $\mathcal{R}_\text{direct}$, and a CoT response, $\mathcal{R}_\text{CoT}$. It then utilizes a weighted average of the cross-modal representation similarity, $\mathcal{S}$, and the model confidence, $\mathcal{C}$, to determine the final answer. 
Specifically, $\mathcal{S} \in [0,1]$ is the cosine similarity between image and text representations after the LLM, while $\mathcal{C} \in (0, 1]$ is the normalized perplexity of the model and formally defined as 
\begin{equation}
\begin{aligned} 
% \mathcal{C} = 2^{\displaystyle\frac{\sum_{t=1}^{T}\log P(w_t|w_1,w_2,\dots,w_{t-1})}{T}},
\mathcal{C} = \exp\left(\frac{\sum_{t=1}^{T}\log P(w_t|w_1,w_2,\dots,w_{t-1})}{T}\right),
\end{aligned}
\label{eq:perplexity}
\end{equation}
where $T$ denotes the number of tokens in the generated text and $P(w_t|w_1,w_2,\dots, w_{t-1})$ is the conditional probability of generating the $t$-th token.

\section{Experiment}
\label{sec:exp}

\begin{algorithm}[!t]
\small
\caption{\small Inference-Time Self-Verification}
\label{alg:algorithm1}
\algorithmicrequire{
Image: $\mathcal{I}$; Question: $\mathcal{Q}$; Task Prompts: $\{\mathcal{P}_\text{direct}, \mathcal{P}_\text{CoT}\}$; \newline
\textcolor{white}{\textbf{Input:}} Averaging Weight: $\alpha$. 
\newline 
}
\algorithmicensure{ 
Answer: $\mathcal{A}$. 
}
\begin{algorithmic}[1]
\State \textcolor{gray}{\textit{\# Computing Similarity and Confidence during inference}}
\State $\mathcal{R}_\text{direct}, \mathcal{S}_\text{direct}, \mathcal{C}_\text{direct} \leftarrow$ $\Ours (\mathcal{I}, \mathcal{Q}, \mathcal{P}_\text{direct})$; 
\State $\mathcal{R}_\text{CoT}, \mathcal{S}_\text{CoT}, \mathcal{C}_\text{CoT} \leftarrow$ $\Ours (\mathcal{I}, \mathcal{Q}, \mathcal{P}_\text{CoT})$; 

\State \textcolor{gray}{\textit{\# Determining the final answer $\mathcal{A}$}}
\State $\mathcal{A}_\text{direct} \gets$ Extracting an answer from $\mathcal{R}_\text{direct}$;
\State $\mathcal{A}_\text{CoT} \gets$ Extracting an answer from $\mathcal{R}_\text{CoT}$;

\If{$\mathcal{A}_\text{CoT} = \mathcal{A}_\text{direct}$} 
\State \Return $\mathcal{A}_\text{CoT}$
\Else
\State $\mathcal{SC}_\text{direct} \gets (1-\alpha)\mathcal{S}_\text{direct} + \alpha\mathcal{C}_\text{direct}$; 
\State $\mathcal{SC}_\text{CoT} \gets (1-\alpha)\mathcal{S}_\text{CoT} + \alpha\mathcal{C}_\text{CoT}$; 
\If{$\mathcal{SC}_\text{CoT} >= \mathcal{SC}_\text{direct}$} 
\State \Return $\mathcal{A}_\text{CoT}$
\Else
\State \Return $\mathcal{A}_\text{direct}$
\EndIf
\EndIf
\end{algorithmic}
\end{algorithm}

\begin{table*}[!t]
\centering
\footnotesize
\setlength{\tabcolsep}{1.mm}{
\begin{tabular}{r|cccc|ccccccc} 
\toprule
&\multicolumn{4}{c|}{\multirow{1}{*}{\textit{Problem Solving}}} 
&\multicolumn{5}{c}{\multirow{1}{*}{\textit{Mathematical Reasoning}}} 
\\
\multicolumn{1}{c|}{\multirow{-2}{*}{{MLLMs}}} 
&MMStar
&MMMU
&SQA-IMG%$^\text{I}$
&AI2D 

&MathVista
&MathVerse
&WeMath
&MathVision
&DynaMath
\\
\midrule

LLaVA-v1.5-7B~\cite{liu2024improvedllava}
&33.1 &35.7 &69.2 &55.5 
&25.5 &4.3 &7.0 &11.4 &1.4 
\\ 
Janus-Pro-7B~\cite{chen2025janus}
&46.5 &41.6 &83.2 &68.1
&42.5 &15.9 &9.7 &14.7 &4.0
\\ 
Molmo-7B-D~\cite{deitke2024molmo}  
&54.4 &48.7 &92.2 &79.6 
&48.7 &4.2 &- &16.2 &12.6 
\\ 
GLM-4v-9B~\cite{glm2024chatglm}
&54.8 &46.9 &\underline{96.7} &71.2
&52.2 &15.9 &11.8 &15.0 &8.6
\\ 
MiniCPM-V-2.6-7B~\cite{yao2024minicpm}
&57.5 &49.8 &\underline{96.7} &82.1 
&60.8 &17.6 &- &18.4 &9.8 
\\ 
URSA-8B~\cite{luo2025ursa}
&- &- &- &-
&58.8 &31.0 &32.8 &28.7 &13.2 
\\ 
InternVL2.5-4B-MPO~\cite{wang2024enhancing}
&- &- &- &-
&64.1 &26.0 &- &22.5 &10.0
\\ 
VITA-v1.5-7B~\cite{fu2025vita15}
&60.2 &52.6 &95.8 &79.2 
&66.2 &23.4 &19.4 &19.5 &9.6 
\\ 
POINTS1.5-7B~\cite{liu2024points1}
&61.1 &53.8 &95.0 &81.4
&66.4 &26.6 &24.6 &22.0 &14.2 
\\ 
Ovis2-4B~\cite{lu2024ovis} 
&61.6 &49.0 &94.0 &85.7
&69.6 &38.5 &16.9 &21.5 &18.0 
\\ 
InternVL2.5-8B~\cite{chen2024expanding}
&63.2 &56.2 &- &84.6 
&64.5 &22.8 &23.5 &17.0 &9.4
\\ 
MiniCPM-o-2.6-7B~\cite{yao2024minicpm}
&63.3 &50.9 &\textbf{98.8} &\underline{86.1} 
&\textbf{73.3} &35.0 &25.2 &21.7 &10.4 
\\ 
Qwen2.5-VL-7B~\cite{bai2025qwen2.5-vl}
&64.1 &56.2 &88.3 &84.1
&65.8 &31.5 &36.4 &25.8 &14.8 
\\ 
Ovis2-8B~\cite{lu2024ovis} 
&\underline{64.6} &\underline{57.4} &95.1 &\textbf{86.6} 
&71.8 &\textbf{42.3} &27.2 &25.9 &20.4 
\\ 
\midrule
\rowcolor{gray!20}
\multicolumn{10}{c}{$\blacktriangledown$~\textit{base LLM: Llama3-8B-Instruct}}
\\ 
VILA1.5-8B~\cite{lin2024vila}
&39.7 &37.4 &73.2 &58.8 
&37.4 &- &- &- &- 
\\ 
Mantis-8B~\cite{jiang2024mantis} 
&41.3 &41.1 &75.5 &60.4 
&32.7 &- &- &- &- 
\\ 
Slime-8B~\cite{zhang2024beyond}
&43.5 &38.8 &78.0 &68.5 
&41.8 &22.9 &- &- &- 
\\
LLaVA-NeXT-8B~\cite{liu2024llavanext}
&43.9 &43.1 &73.1 &72.8 
&37.7 &- &- &- &- 
\\
Idefics3-8B~\cite{laurenccon2024building}
&55.0 &46.6 &91.3 &76.5 
&58.7 &- &- &- &- 
\\ 
MiniCPM-V-2.5-8B~\cite{yao2024minicpm}
&51.8 &45.8 &89.2 &78.4 
&54.5 &- &- &- &- 
\\ 
Bunny-8B~\cite{he2024efficient}
&45.4 &43.4 &79.1 &69.4 
&35.2 &- &- &- &-  
\\ 
Ovis1.5-8B~\cite{lu2024ovis} 
&57.3 &48.3 &88.8 &82.5 
&63.0 &- &- &- &- 
\\ 
Cambrian-8B~\cite{tong2024cambrian}
&- &42.7 &80.4 &73.0 
&49.0 &- &- &- &- 
\\
Eagle-X5-8B~\cite{shi2024eagle}
&- &43.8 &84.3 &76.1 
&52.7 &- &- &- &- 
\\ 
% \midrule 
\rowcolor{ggreen!20}\textbf{\Ours-base-8B} 
&62.4 &\underline{57.4} &93.2 &82.8 
&64.8 &34.8 &\underline{54.0} &\underline{26.8} &\underline{24.5}
\\
\rowcolor{ggreen!20}\textbf{\Ours-o1-8B}
&\textbf{65.2} &\textbf{59.7} &93.8 &85.0
&\underline{72.0} &\underline{40.1} &\textbf{59.8} &\textbf{30.1} &\textbf{33.2}
\\ 
\bottomrule
\end{tabular}
}
\vspace{-2mm}
\caption{\textbf{Comparison with state-of-the-art MLLMs on multimodal reasoning benchmarks.}
The best and runner-up results in each section are highlighted in \textbf{bold} and \underline{underlined}, respectively. 
The performance of compared MLLMs is mainly derived from VLMEvalKit. 
}
\label{tab:comp_reasoning}
% \vspace{-2mm}
\end{table*}

\subsection{Experimental Setups}

\vspace{2pt}
\pgraph{Benchmarks}
In this work, we primarily evaluate the reasoning capabilities of \Ours across 9 multimodal reasoning benchmarks, including MMStar~\cite{chen2024we}, MMMU~\cite{yue2023mmmu}, SQA-IMG~\cite{lu2022learn}, AI2D~\cite{kembhavi2016diagram}, MathVista~\cite{lu2023mathvista}, MathVerse~\cite{zhang2024mathverse}, WeMath~\cite{qiao2024we}, MathVision~\cite{wang2024measuring}, and DynaMath~\cite{zou2024dynamath}. These benchmarks encompass tasks involving science problem solving and mathematical reasoning. 
For a comprehensive evaluation of general multimodal abilities, we employ 4 representative benchmarks: SEED-IMG~\cite{li2023seed}, MMT-Val~\cite{ying2024mmt}, RWQA~\cite{xai2024grok}, and BLINK~\cite{fu2024blink}. 
We report the official performance metric (\ie, accuracy) computed using an open-source evaluation toolkit of MLLMs (\ie, VLMEvalKit~\cite{duan2024vlmevalkit}) for all the aforementioned benchmarks.

\vspace{2pt}
\pgraph{Implementation Details}
We set the sequence length $N_\emph{p}$ in \mmconnector to 24. In training phrases, the maximum sequence length of LLMs is set to 2,048. During inference, the weight factor $\alpha$ in \Cref{alg:algorithm1} is set to 0.7, and the maximum length of generated tokens is limited to 1,024. For \Ours-base, we set the $\mathrm{temperature}$ of LLM generation to 1.0, whereas for \Ours-o1, we set the $\mathrm{temperature}$ and $\mathrm{top_p}$ to 0.4 and 0.9, respectively. All training and inference are conducted on 8 NVIDIA A800 (80G) GPUs. 
% \cref{sec:app_training} provides detailed training setups and costs. 
Detailed training setups and costs are provided in the Supplementary Material.

% ****************************************************************************
\subsection{Main Results}

\vspace{2pt}
\pgraph{Multimodal Reasoning Benchmarks} 
In \cref{tab:comp_reasoning}, we compare the proposed \Ours with open-source MLLMs with on-par parameter scales and customized MLLMs using the same base LLM. 
For problem-solving tasks, Corvid-o1-8B outperforms other custom MLLMs on all benchmarks. Notably, Corvid-o1-8B achieves scores of 65.2 and 59.7 on MMStar and MMMU respectively, surpassing prior best-performing models. For mathematical reasoning, Corvid-o1-8B also performs well across all benchmarks and achieves the highest scores on WeMath (59.8), MathVision (30.1), and DynaMath (33.2). These results confirm Corvid-o1's advanced multimodal reasoning, suggesting that improving the CoT reasoning capability of MLLMs is a feasible solution to handle challenging multimodal tasks.

\vspace{2pt}
\pgraph{Comprehensive Benchmarks} 
\cref{tab:comp_understand} presents a comparison across comprehensive benchmarks, revealing that our models outperform other Llama3-8B-Instruct-based counterparts by a large margin and achieve competitive performance compared to leading MLLMs. 
In particular, on RWQA, \Ours-base-8B and \Ours-o1-8B surpass the current state-of-the-art by 4.5 and 3.9 points, respectively. 
On the challenging vision-centric benchmark BLINK, \Ours-o1-8B attains a top score of 57.9, outperforming the runner-up by 1.4 points. 
Overall, these results demonstrate the robust general multimodal abilities of \Ours.

% ****************************************************************************
\subsection{Ablation Studies}

\vspace{2pt}
\pgraph{Effectiveness of \mmconnector} 
As shown in~\cref{tab:abl_gatemixer}, we first ablate two critical components -- the gate attention mechanism ($\mathcal{M}_\text{GA}$) and prefix token embeddings ($\hv_p$) -- along with a training objective ($\mathcal{L}_\text{CReg}$) of \mmconnector to validate its effectiveness. 
Without $\mathcal{M}_\text{GA}$ and $\mathcal{L}_\text{CReg}$ alignment training, Corvid's performance declined across all nine benchmarks. Similarly, removing $\hv_p$ resulted in performance degradation on seven benchmarks. 
Additionally, we compare the performance of \mmconnector with two MLP-based typical architectures, FC\_GELU\_FC and 2$\times$FC\_GELU\_FC, to further assess its effectiveness. The results in \cref{tab:abl_gatemixer} show that \mmconnector outperforms these MLP-based connectors by 1.6\% and 2.3\% on average, respectively, highlighting the superiority of our connector. 
Overall, these findings confirm the effectiveness of GateMixer configurations.

% **********************************************************

\begin{table}[!t]
\centering
\footnotesize
\setlength{\tabcolsep}{.8mm}{
\begin{tabular}{r|cccc} 
\toprule
\multicolumn{1}{c|}{\multirow{1}{*}{{MLLMs}}} 
&SEED-IMG
&MMT-Val
&RWQA
&BLINK
\\
\midrule
Monkey-Chat-7B~\cite{li2024monkey}
&68.9 &53.3 &52.4 &47.1
\\ 
DeepSeek-VL-7B~\cite{lu2024deepseek}
&70.1 &53.5 &54.2 &40.9 
\\ 
Molmo-7B-D~\cite{deitke2024molmo}  
&74.1 &56.8 &68.2 &46.1
\\ 
VITA-v1.5-7B~\cite{fu2025vita15}
&74.1 &59.5 &66.9 &45.0
\\ 
POINTS1.5-7B~\cite{liu2024points1}
&75.1 &61.8 &67.5 &44.0
\\ 
MiniCPM-V-2.6-7B~\cite{yao2024minicpm}
&74.0 &60.8 &65.0 &55.2
\\ 
Qwen2.5-VL-7B~\cite{bai2025qwen2.5-vl}
&76.9 &62.7 &68.8 &56.1
\\ 
InternVL2.5-8B-MPO~\cite{wang2024enhancing}
&76.8 &62.5 &68.8 &\underline{56.6}
\\
Ovis2-8B~\cite{lu2024ovis} 
&\textbf{77.2} &\textbf{66.6} &72.5 &54.3 
\\ 

\midrule
\rowcolor{gray!20}
\multicolumn{5}{c}{$\blacktriangledown$~\textit{base LLM: Llama3-8B-Instruct}}
\\ 
VILA1.5-8B~\cite{lin2024vila}
&65.0 &48.7 &43.4 &39.5 
\\ 
Slime-8B~\cite{zhang2024beyond}
&69.8 &50.2 &58.0 &38.8 
\\
Mantis-8B~\cite{jiang2024mantis}
&71.2 &54.3 &59.5 &50.1 
\\ 
LLaVA-NeXT-8B~\cite{liu2024llavanext}
&72.5 &53.1 &58.4 &43.5 
\\
Bunny-8B~\cite{he2024efficient}
&73.5 &54.8 &60.4 &41.6 
\\ 
Eagle-X5-7B~\cite{shi2024eagle}
&73.6 &52.6 &63.8 &22.4 
\\ 
Ovis1.5-8B~\cite{lu2024ovis} 
&75.4 &60.7 &64.2 &39.8 
\\ 
Idefics3-8B~\cite{laurenccon2024building}
&73.8 &58.4 &62.6 &50.3
\\ 
% \midrule
\rowcolor{ggreen!20}
\textbf{\Ours-base-8B} 
&76.4 &62.5 &\textbf{77.0} &55.7 
\\
\rowcolor{ggreen!20}
\textbf{\Ours-o1-8B} 
&\underline{77.0} &\underline{65.9} &\underline{76.4} &\textbf{57.9}
\\
\bottomrule
\end{tabular}
}
\vspace{-2mm}
\caption{\textbf{Results on comprehensive multimodal benchmarks.}
% All results are derived from VLMEvalKit. 
}
\label{tab:comp_understand}
\vspace{-2mm}
\end{table}

% *******************************************************************
\begin{table*}[!th]
\footnotesize  
\centering 
\setlength{\tabcolsep}{1.4mm}{
\begin{tabular}{l|c|ccccccccc}
\toprule
{\quad Connector}
&\textbf{Average}
&MMStar
&MMMU
&SQA-IMG
&AI2D
&MathVista
&MathVerse
&WeMath
&MathVision
&DynaMath
\\ 
\midrule
\rowcolor{gray!20}
FC\_GELU\_FC 
&54.0
&61.8 &54.9 &92.5 &80.7
&62.1 &32.9 &49.9 &24.2 &\textbf{26.5}
\\ 
\rowcolor{gray!20}
2$\times$FC\_GELU\_FC
&53.7
&61.6 &54.6 &92.2 &82.1
&63.2 &32.6 &51.2 &23.6 &22.3
\\ 
\midrule
% \rowcolor{ggreen!20}
{\mmconnector} 
&\textbf{55.6}
&\textbf{62.4} &57.4 &\textbf{93.2} &\textbf{82.8}  
&\textbf{64.8} &\textbf{34.8} &54.0 &\textbf{26.8} &24.5
\\
{\qquad w/o $\mathcal{L}_\text{CReg}$}
&55.0
&60.1 &57.0 &91.7 &82.0
&64.3 &32.9 &57.1 &26.0 &23.9
\\
{\qquad w/o $\hv_{\emph{p}}$}
&54.8
&60.9 &\textbf{58.8} &92.9 &82.5 
&58.1 &32.6 &\textbf{60.0} &24.1 &23.6
\\
{\qquad w/o GA}
&54.1
&58.1 &57.1 &89.9 &82.1 
&64.1 &29.8 &56.5 &24.7 &24.1
\\
\bottomrule
\end{tabular}
}
\vspace{-2mm}
\caption{\textbf{Ablation study on \mmconnector.} 
\mmconnector improves feature integration via gate attention mechanism ($\mathcal{M}_\text{GA}$) and prefix token embeddings ($\hv_{\emph{p}}$). $\mathcal{L}_\text{CReg}$ facilitates cross-modal alignment during pretraining. 
2$\times$FC\_GELU\_FC: two separate FC layers followed by a GELU activation and a FC layer. 
FC\_GELU\_FC: one FC layer followed by a GELU activation and a FC layer. 
}
\label{tab:abl_gatemixer}
% \vspace{-1mm}
\end{table*}

% *******************************************************************
\begin{table*}[!th]
\footnotesize  
\centering 
\setlength{\tabcolsep}{1.4mm}{
\begin{tabular}{l|c|ccccccccc}
\toprule
\multicolumn{1}{c|}{SFT Data}
&\textbf{Average}
&MMStar
&MMMU
&SQA-IMG
&AI2D
&MathVista
&MathVerse
&WeMath
&MathVision
&DynaMath
\\ 
\midrule
\rowcolor{gray!20}
LLaVA-665K
&44.1
&51.0	&49.5	&89.6	&66.7	
&43.8	&26.7	&35.7	&16.7	&16.8
\\ 
% Bunny-695K
% \\ 
\rowcolor{gray!20}
Cambrian-1M$^\dagger$
&51.9
&57.7 &60.7	&88.8 &77.0	
&57.0 &31.1	&46.8 &25.0	&22.7
\\ 
\rowcolor{gray!20}
Ovis-1M$^\dagger$
&53.8 
&61.3 &\textbf{57.7} &93.0 &81.6
&61.3 &32.9 &49.9 &24.1 &21.9
\\ 

\midrule
\Ours-1M
&\textbf{55.6}
&\textbf{62.4} &{57.4} &\textbf{93.2} &\textbf{82.8}  
&\textbf{64.8} &\textbf{34.8} &\textbf{54.0} &\textbf{26.8} &\textbf{24.5}
\\ 
$\mathcal{D}_\text{raw rationale}$-1M
&51.2
&55.0 &55.1 &90.7 &81.9 
&54.5 &28.8 &49.0 &23.9 &21.9 
\\
$\mathcal{D}_\text{direct}$-1M
&45.7
&49.8 &44.6 &85.7 &79.1 
&50.9 &24.3 &44.4 &13.6 &18.5 
\\
\bottomrule
\end{tabular}
}
\vspace{-2mm}
\caption{\textbf{Effect of high-quality CoT data.} 
Cambrian-1M$^\dagger$ and Ovis-1M$^\dagger$ are sampled from Cambrian-10M~\cite{tong2024cambrian} and Ovis-dataset~\cite{lu2024ovis}. 
}
\label{tab:abl_sft_data}
% \vspace{-1mm}
\end{table*}

% *******************************************************************
\begin{table*}[!th]
\footnotesize  
\centering 
\setlength{\tabcolsep}{1.4mm}{
\begin{tabular}{c|l|c|ccccccccc}
\toprule
MLLMs
&\multicolumn{1}{c|}{Strategy}
&\textbf{Average}
&MMStar
&MMMU
&SQA-IMG
&AI2D
&MathVista
&MathVerse
&WeMath
&MathVision
&DynaMath
\\ 
\midrule
&\textsc{Infer}$_\text{<direct>}$
&44.4
&54.1 &45.9 &87.3 &78.3 
&44.3 &22.8 &31.6 &15.4 &20.3 
\\
&\textsc{Infer}$_\text{<CoT>}$
&49.7
&54.3 &52.8 &89.4 &67.7 
&57.6 &32.5 &46.7 &22.4 &23.9 
\\
\multirow{-3}{*}{\Ours-base}
&\textsc{Infer}$_\text{<SV>}$
&\textbf{55.6}
&\textbf{62.4} &\textbf{57.4} &\textbf{93.2} &\textbf{82.8}  
&\textbf{64.8} &\textbf{34.8} &\textbf{54.0} &\textbf{26.8} &\textbf{24.5}
\\ 
\midrule
&\textsc{Infer}$_\text{<direct>}$
&48.9
&54.5 &47.6 &86.8 &79.5 
&49.2 &32.2 &41.8 &24.1 &24.7
\\ 
&\textsc{Infer}$_\text{<CoT>}$
&56.2
&61.1 &55.7 &91.6 &77.5
&67.2 &36.5 &57.1 &27.6 &31.9
\\ 
\multirow{-3}{*}{\Ours-o1}
&\textsc{Infer}$_\text{<SV>}$
&\textbf{59.9}
&\textbf{65.2} &\textbf{59.7} &\textbf{93.8} &\textbf{85.0}
&\textbf{72.0} &\textbf{40.1} &\textbf{59.8} &\textbf{30.1} &\textbf{33.2}
\\ 
\bottomrule
\end{tabular}
}
\vspace{-2mm}
\caption{
\textbf{Effectiveness of the proposed self-verification strategy.} 
Our inference strategy (\textsc{Infer}$_\text{<SV>}$) showcases significant performance improvements over \underline{direct inference} (\textsc{Infer}$_\text{<direct>}$) and \underline{inference with CoT reasoning} (\textsc{Infer}$_\text{<CoT>}$). 
}
\label{tab:abl_infer}
% \vspace{-2mm}
\end{table*}

% **********************************************************

\vspace{2pt}
\pgraph{Effect of High-Quality CoT Data}
We argue that high-quality CoT-formatted data is essential for improving existing MLLMs towards CoT reasoning.
To further investigate this, we compare our final SFT dataset (\Ours-1M) with two alternative datasets: 
\emph{(i)} $\mathcal{D}_\text{direct}$-1M, deleting CoTs and keeping only final answers in \Ours-1M. 
\emph{(ii)} $\mathcal{D}_\text{raw rationale}$-1M, replacing our high-quality CoTs with the corresponding raw rationales from MCoT-Instruct's source datasets. 
The performance of models fine-tuned on these datasets is presented in~\cref{tab:abl_sft_data}, showing that \Ours-1M outperforms \emph{(i)} and \emph{(ii)} across all benchmarks, underscoring the importance of detailed and standardized CoT data.
Additionally, we compare \Ours-1M with three commonly used SFT datasets in \cref{tab:abl_sft_data}, showing that \Ours-1M surpasses them by 11.5\%, 8.3\%, and 6.7\% on average, respectively. 
These results further emphasize the significance of high-quality CoT-formatted data in improving CoT reasoning capabilities.

% **********************************************************
\begin{figure}[!t]
\centering
\includegraphics[width=0.98\linewidth]{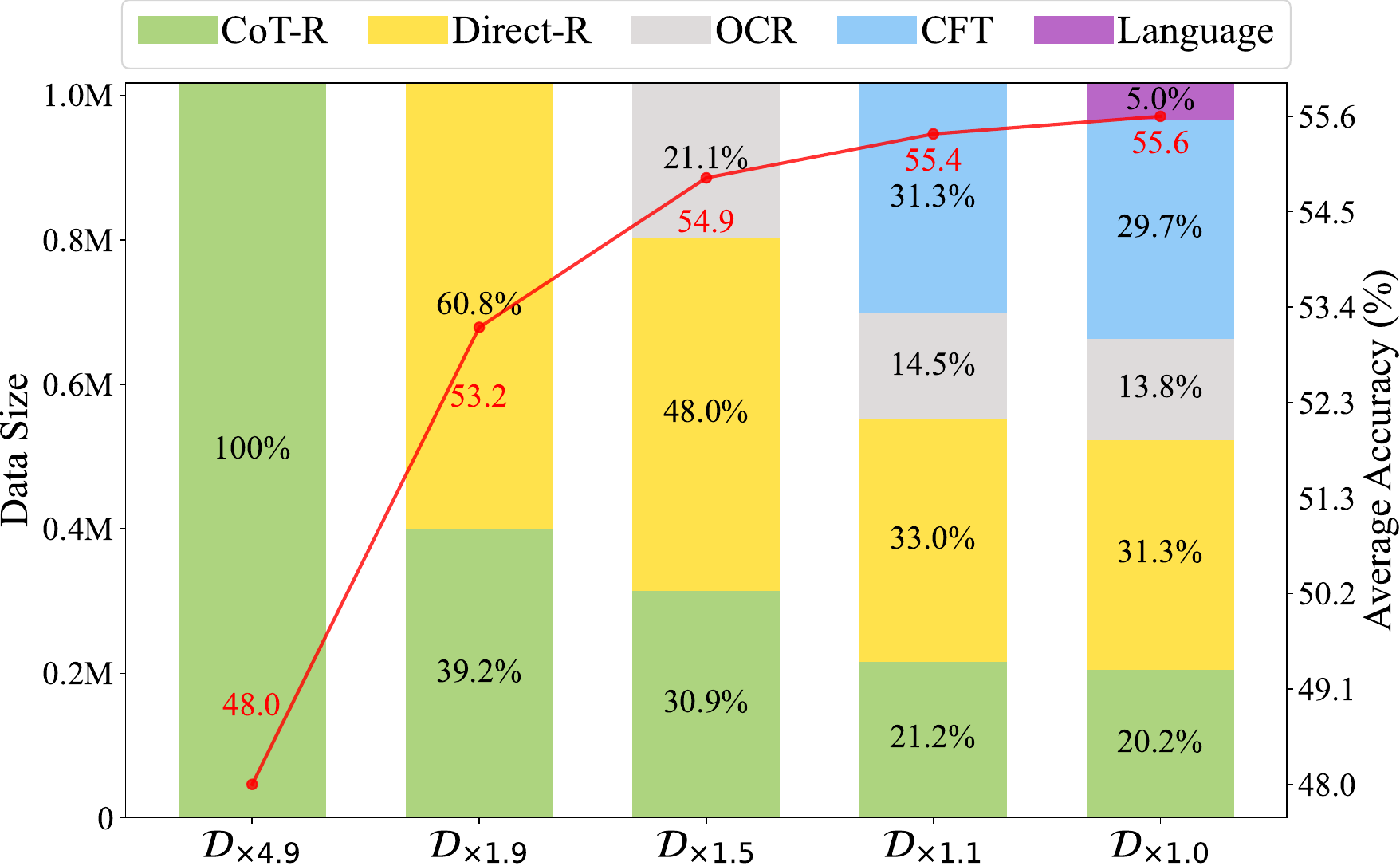}
\vspace{-2mm}
\caption{
\textbf{Effect of SFT data diversity.}
The number in the subscript of $\mathcal{D}$ indicates the data repetition rate. $\mathcal{D}_{\times 1.0}$ i.e. Covid-1M.
}
\label{fig:sft_data_type}
\vspace{-4mm}
\end{figure}
% **********************************************************

\vspace{2pt}
\pgraph{Effect of SFT Data Diversity}
In \cref{fig:sft_data_type}, we further explore the impact of SFT data diversity on model performance with a controlled data size of 1M. 
As the dataset transitions from $\mathcal{D}_{\times 4.9}$ (with only CoT reasoning data) to \Ours-1M, the average accuracy steadily increases from 48.0 to 55.6, indicating that incorporating different diverse data types (\eg, direct reasoning and OCR data) facilitates CoT reasoning performance. 
These results highlight the importance of diverse and well-structured SFT data in improving model reasoning capabilities under a fixed data volume.

% *******************************************************************
\begin{table*}[!th]
\footnotesize  
\centering 
\setlength{\tabcolsep}{1.4mm}{
\begin{tabular}{cc|c|ccccccccc}
\toprule
SigLIP &ConvNeXt
&\textbf{Average}
&MMStar
&MMMU
&SQA-IMG
&AI2D
&MathVista
&MathVerse
&WeMath
&MathVision
&DynaMath
\\ 
\midrule
\checkmark &
&54.7
&\textbf{63.2} &\textbf{58.9} &\textbf{94.2} &81.6 
&55.5 &34.1 &\textbf{55.0} &25.0 &\textbf{24.7} 
\\ 
&\checkmark
&52.3
&59.1 &55.6 &90.4 &80.4 
&55.4 &29.7 &50.6 &25.4 &24.2
\\
\checkmark &\checkmark
&\textbf{55.6}
&62.4 &57.4 &93.2 &\textbf{82.8}  
&\textbf{64.8} &\textbf{34.8} &54.0 &\textbf{26.8} &24.5
\\ 
\bottomrule
\end{tabular}
}
\vspace{-2mm}
\caption{\textbf{Effect of the hybrid vision encoder} that combines SigLIP and ConvNeXt for more informative visual representation. 
}
\label{tab:abl_hybrid_ve}
% \vspace{-2mm}
\end{table*}

\begin{table*}[!t]
\footnotesize
\centering
\setlength{\tabcolsep}{1.4mm}{
\begin{tabular}{l|ccccc|ccccc|c} 
\toprule
\multicolumn{1}{c|}{\multirow{2}{*}{{MLLMs}}}
&\multicolumn{5}{c|}{\textit{Similarity-based Metrics}} 
&\multicolumn{5}{c|}{\textit{LLM-based Offline Evaluator}}
&\multicolumn{1}{c}{{GPT-Scorer}} 
\\
&\multicolumn{1}{l}{{{BLUE-1}}}
&\multicolumn{1}{l}{{{BLUE-4}}}
&\multicolumn{1}{l}{{{ROUGE-L}}}
&\multicolumn{1}{l}{{{METEOR}}}
&\multicolumn{1}{l|}{{{BERTScore}}}
&\multicolumn{1}{l}{{{Fact-C}}}
&\multicolumn{1}{l}{{{C-Nat.}}}
&\multicolumn{1}{l}{{{C-Coh.}}}
&\multicolumn{1}{l}{{{C-Eng.}}}
&\multicolumn{1}{l|}{{{C-Gnd.}}}
&\multicolumn{1}{c}{{Overall}}
\\
\midrule
\Ours-base
&0.362
&0.162
&0.445
&0.453 
&0.763

&0.792
&0.829
&0.998
&7.137
&0.990 

&0.940
\\ 
\Ours-o1
&0.541
&0.396
&0.616
&0.664
&0.894

&0.847
&0.894
&0.999
&9.491
&0.975

&0.990
\\

\bottomrule
\end{tabular}
}
\vspace{-2mm}
\caption{\textbf{Automatic quality evaluation of CoT responses on SQA-IMG} with human-provided CoTs as references.  
We use the following abbreviations: Fact-C for factual consistency, and C-Nat./Coh./Eng./Gnd. for naturalness, coherence, engagingness, and groundedness in conversation. 
Except for C-Eng. $\in [0, +\infty)$, all other metrics range from 0 to 1. 
}
\label{tab:result_cot_quality}
% \vspace{-2mm}
\end{table*}

\vspace{2pt}
\pgraph{Effectiveness of Self-Verification Strategy (\textsc{Infer}$_\text{<SV>}$)}
This strategy is a inference-time scaling technique designed to mitigate \textit{over-reasoning} and \textit{under-reasoning} issues. 
To evaluate its effectiveness, we compare it against two alternative settings: 
\textsc{Infer}$_\text{<CoT>}$, which formats testing samples as hard instances and performs CoT reasoning before answering, and \textsc{Infer}$_\text{<direct>}$, which treats testing samples as easy instances and performs direct inference. 
The results in \cref{tab:abl_infer} demonstrate that across both \Ours-base and \Ours-o1, \textsc{Infer}$_\text{<SV>}$ consistently outperforms the two alternatives. In \Ours-base, it achieves an average score of 55.6, surpassing \textsc{Infer}$_\text{<direct>}$ (44.4) by 11.2 points and \textsc{Infer}$_\text{<CoT>}$ (49.7) by 5.9 points. A similar improvement is observed in \Ours-o1, where \textsc{Infer}$_\text{<SV>}$ exceeds them by 11.0 and 3.7 points, respectively. 
These performance gains showcase the effectiveness of our self-verification strategy.

% ****************************************************************************
\subsection{Analysis and Discussion}

\vspace{2pt}
\noindent\textbf{Is Hybrid Vision Encoder Necessary?\,} 
To investigate this, we conduct the ablation study presented in~\cref{tab:abl_hybrid_ve}. 
The results indicate that leveraging both encoders yields the highest average performance (55.6) across multiple benchmarks, outperforming the use of SigLIP (54.7) or ConvNeXt (52.3) alone. 
Notably, the hybrid encoder achieves the best scores in AI2D (82.8), MathVista (64.8), MathVerse (34.8), and MathVision (26.8), demonstrating enhanced capabilities in both scientific and mathematical reasoning. 
These findings confirm the necessity of enriching visual representations using a hybrid vision encoder.

\vspace{2pt}
\pgraph{CoT Quality Evaluation} 
We utilize three types of evaluation methods to assess the quality of \Ours responses:  
\emph{(i)} similarity-based metrics (BLEU-1\&4~\cite{papineni2002bleu}, ROUGE-L~\cite{lin2004rouge}, METEOR~\cite{banerjee2005meteor}, and BERTScore~\cite{nils2020sentence}) to measure the coverage and similarity between generated and reference CoTs. 
\emph{(ii)} an LLM-based offline evaluator, \textsc{UniEval}~\cite{zhong4towards}, to evaluate the factual consistency, naturalness, coherence, engagingness, and groundedness of generated CoTs. 
\emph{(iii)} a GPT-based online evaluator to assess CoTs across faithfulness, relevance, and completeness, providing an overall score. 
Results in \cref{tab:result_cot_quality} indicate that while \Ours's CoT responses exhibit low coverage compared to human-provided ones, they demonstrates high overall performance, particularly excelling in naturalness, coherence, and engagingness, thereby indirectly validating their quality.

\begin{figure}[!t]
\centering
\includegraphics[width=0.99\linewidth]{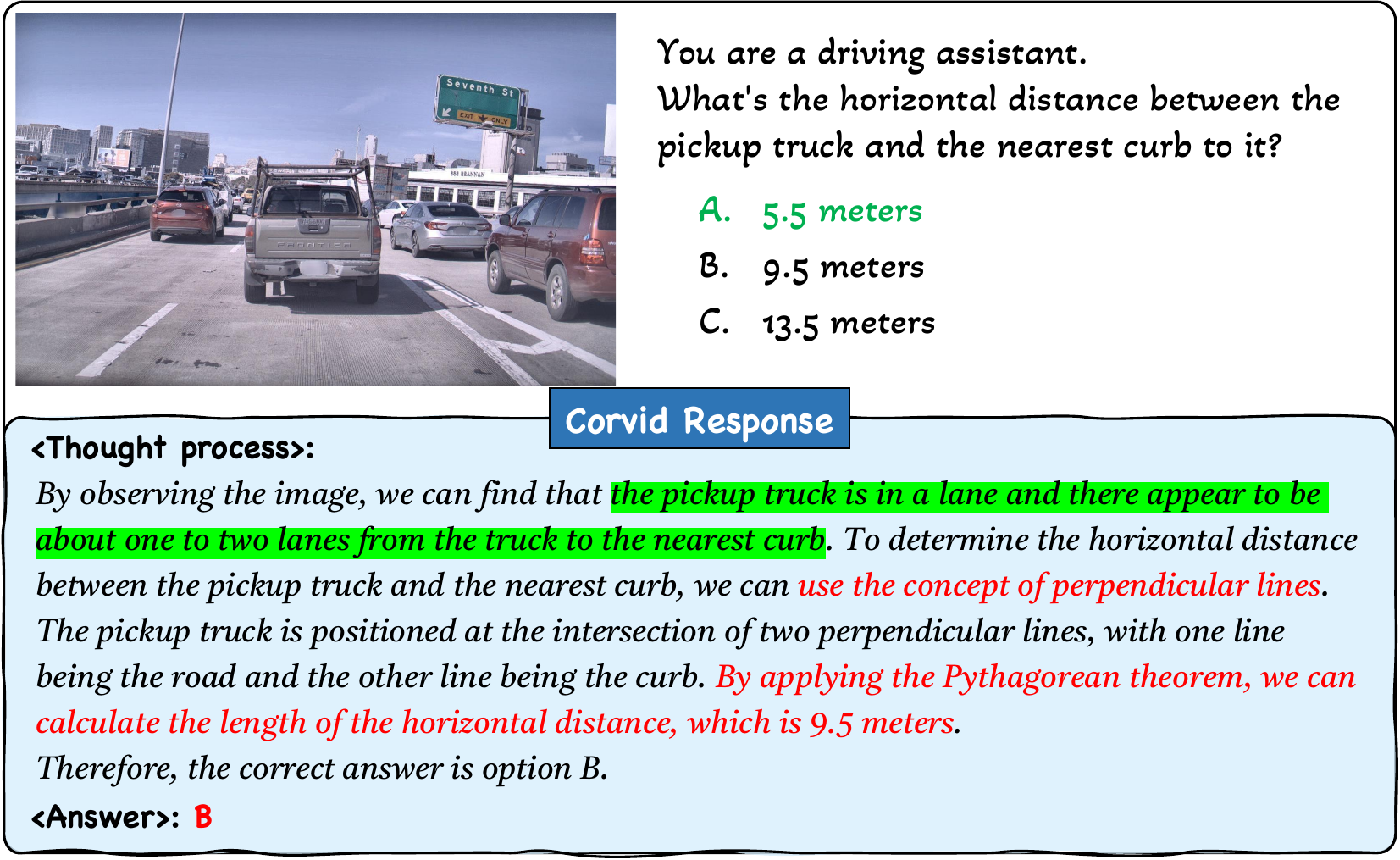}
\vspace{-2mm}
\caption{
\textbf{A failure case} where \Ours fails to reason effectively due to a lack of world commonsense. 
}
% \vspace{-2mm}
\label{fig:app_case}
\end{figure}

\vspace{2pt}
\pgraph{Case Study on Corvid's Limitations} 
To analyze Corvid's reasoning limitations, we conduct an in-depth analysis of failure cases and observe that it struggles with specific types of reasoning problems, particularly those \textit{requiring world knowledge or commonsense}. For example, as illustrated in \cref{fig:app_case}, while Corvid accurately locates the positions of the pickup truck and the nearest curb, it fails to reason correctly due to a lack of world commonsense (\ie, typical lane widths on urban roads or highways range from about 3.5 to 4.5 meters). Therefore, more techniques, such as group relative policy optimization~\cite{deepseek2025deepseek} and retrieval-augmented generation~\cite{chen2024mllm}, need to be explored to further improve the reasoning capabilities of MLLMs.

% *****************************************************************
% *****************************************************************

%%%%%%%%%%%%%%%%%%%%%%%%%%%%%%%%%%%%%%%%%%%%%%%%%%%%%%%%%%%
\section{Conclusion}

In this work, we introduce Corvid as the primary exploration into CoT reasoning for MLLMs. Enhanced by critical components such as GateMixer, high-quality CoT data, and an inference-time self-verification strategy, Corvid surpasses existing o1-like MLLMs and state-of-the-art MLLMs with comparable parameter scales across various reasoning and comprehensive benchmarks, showcasing its advanced CoT reasoning capabilities. 
Despite its impressive performance, several challenges persist that merit future investigation, such as knowledge-intensive reasoning (as discussed in the Limitations section) and efficient scaling techniques. 
We believe that ongoing research into o1-like capability will further enhance MLLMs to address more sophisticated tasks.

% \clearpage
\pgraph{Acknowledgement} This work was supported in part by NSFC (62322113, 62376156, 62406189), Shanghai Municipal Science and Technology Major Project (2021SHZDZX0102), and the Fundamental Research Funds for the Central Universities.

{
\small
\bibliographystyle{ieeenat_fullname}
\bibliography{main}
}

% % WARNING: do not forget to delete the supplementary pages from your submission 
\clearpage
\appendix
\setcounter{page}{15}
\setcounter{table}{9}
\setcounter{figure}{4}
\maketitlesupplementary

% ************************************************************ 
\section{Supplementary Training Details}
\label{sec:app_training}

In \cref{tab:app_training_param}, we provide detailed hyperparameter settings and time costs for the three-stage training of \Ours. Unless otherwise specified, all training and inference are conducted using 8 NVIDIA A800 (80G) GPUs by default. 

% ***************************************************************************
\begin{table}[!th]
\centering
\footnotesize 
\vspace{-3mm}
\setlength{\tabcolsep}{1.mm}{
\begin{tabular}{l|c|c|c} 
\toprule 
{Configuration}
&{Stage 1}
&{Stage 2}
&{Stage 3}
\\
\midrule 
% Image resolution
% &384$\times$384 &384$\times$384 &384$\times$384
% \\
Batch size 
&256 &256 &128
\\
Peak learning rate
&1e-3 &2e-5 &2e-6 
\\
Learning rate schedule
&Cosine &Cosine &Cosine 
\\ 
Learning rate warm-up ratio
&0.03 &0.03  &0.03 
\\ 
Weight decay
&0 &0 &0
\\ 
% Grad norm clipping
% &1.0 &1.0
% \\ 
Epoch
&1 &1 &3
\\ 
Optimizer
&AdamW &AdamW &AdamW 
\\ 
Float precision
&bfloat16 &bfloat16 &bfloat16
\\
Deepspeed configuration
&zero2 &zero3 &zero3 
\\ 
Training modules 
&\mmconnector &\mmconnector, LLM &LLM 
\\
Data Size 
&1M &1M &320K
\\ 
Training hours 
&$\sim$11 &$\sim$28 &$\sim$20
\\ 
\bottomrule
\end{tabular}
}
\vspace{-3mm}
\caption{\textbf{Training hyperparameter setting.}}
\label{tab:app_training_param}
\vspace{-3mm}
\end{table}

% *************************************************************
\section{Additional Experiment Results}
\label{sec:app_exp}

\subsection{Comparison with o1-Like MLLMs}

In \cref{tab:bench}, we compare our models against o1-like MLLMs on various benchmarks, including MMStar~\cite{chen2024we}, MMB~\cite{liu2023mmbench}, MMVet~\cite{yu2023mm}, MathVista (MathV)~\cite{lu2023mathvista}, AI2D~\cite{kembhavi2016diagram}, and Hallusion~\cite{guan2024hallusionbench}, using their benchmark metrics computed with official implementations. 
Here, \Ours-o1\textsuperscript{\textdagger}, LLaVA-o1~\cite{xu2024llava}, and LlamaV-o1~\cite{thawakar2025llamav} utilize the same baseline Llama-3.2-11B-Vision-Instruct~\cite{llama32v}. 
Results in the table showcase that Corvid-o1-8B surpasses existing o1-like MLLMs on multiple benchmarks, particularly outperforming llamaV-o1 and Mulberry-o1-7B~\cite{yao2024mulberry} on MathVista by 10.5 and 14.5 points, respectively. Additionally, Corvid-o1 achieves the best overall performance across all benchmarks. These results highlight the effectiveness of Corvid-o1, establishing it as a competitive MLLM that exceeds existing o1-like MLLMs with similar parameter sizes.

\subsection{Additional Evaluation on VRC-Bench}

We additionally evaluate our model on VRC-Bench~\cite{thawakar2025llamav}, which is specifically designed for multimodal step-by-step reasoning tasks. The results in \cref{tab:vrc_bench} show that \Ours-o1 achieves leading accuracy in final answers but exhibits limited performance on reasoning steps. This is because, compared to Llava-CoT and LlamaV-o1, \Ours-o1's reasoning traces do not strictly adhere to the annotated multi-step structure in VRC-Bench. It tends to generate more streamlined and simplified reasoning processes rather than following the predefined step-by-step format. 

\begin{table}[h]
\footnotesize
\centering
\setlength{\tabcolsep}{.55mm}{
\begin{tabular}{lccccc}
\toprule
% \hline 
\multirow{2}{*}{Model}
&Llama-3.2 &Mulberry &LLaVA-o1 &LlamaV-o1 &\Ours-o1
\\
&Vision~\cite{llama32v}
&~\cite{yao2024mulberry}
&~\cite{xu2024llava}
&~\cite{thawakar2025llamav}
&(Ours)
\\ 
\midrule
% \hline 
Final Answer
&48.40 &51.90 &54.09 &{56.49}
&61.90
\\
Steps 
&58.37 &63.86 &66.21 &{68.93}
&63.93
\\
\bottomrule
% \hline 
\end{tabular}
}
\vspace{-2mm}
\caption{\textbf{Comparison with o1-like MLLMs on VRC-Bench.} 
}
\label{tab:vrc_bench}
\vspace{-2mm}
\end{table}

\subsection{Influence of $\alpha$ on Self-Verification}
\label{sec:alpha_influence}

\begin{figure}[!h]
\centering 
\includegraphics[width=0.9\linewidth]{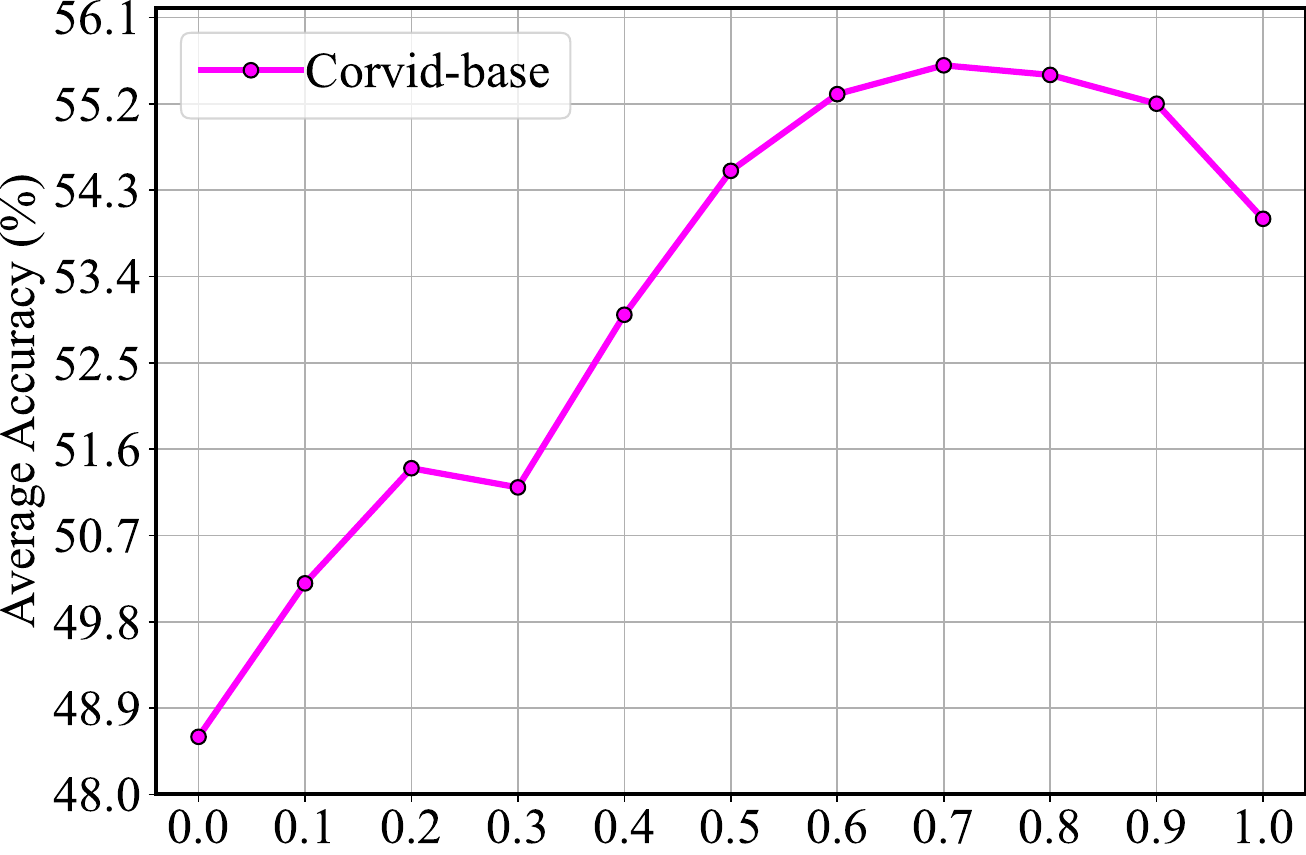}
\vspace{-3mm}
\caption{\textbf{Influence of $\alpha$ on our self-verification strategy.} 
}
\label{fig:acc_alpha}
\vspace{-5mm}
\end{figure}

In the proposed self-verification strategy, $\alpha$ is a weighting factor used to trade-off the cross-modal representation similarity $\mathcal{S}$ and the model confidence $\mathcal{C}$ for the final decision $\mathcal{SC}$. This relationship is formally expressed as:
$$\mathcal{SC} = (1 - \alpha)\mathcal{S} + \alpha \mathcal{C}.$$ 
To analyze the influence of $\alpha$, we conduct ablation studies by varying $\alpha$ from 0.0 to 1.0 with a step size of 0.1. 
\cref{fig:acc_alpha} illustrates the relationship between model performance and the weighting factor $\alpha$ in our self-verification strategy. 
As $\alpha$ increases from 0.0 to 0.7, accuracy rises significantly from 48.6 to a peak of 55.6, demonstrating the advantage of incorporating model confidence into the final answer selection. 
Beyond $\alpha$ = 0.7, performance gradually declines, suggesting that overweighting confidence relative to cross-modal similarity degrades effectiveness. 
The optimal value ($\alpha$ = 0.7) indicates that while both components contribute meaningfully to verification performance, a configuration that slightly prioritizes confidence yields superior results.

% ******************************************************************
% ******************************************************************

% *********************************************************
\begin{figure*}[!t]
\begin{subfigure}{\linewidth}
\centering
\includegraphics[width=0.99\linewidth]{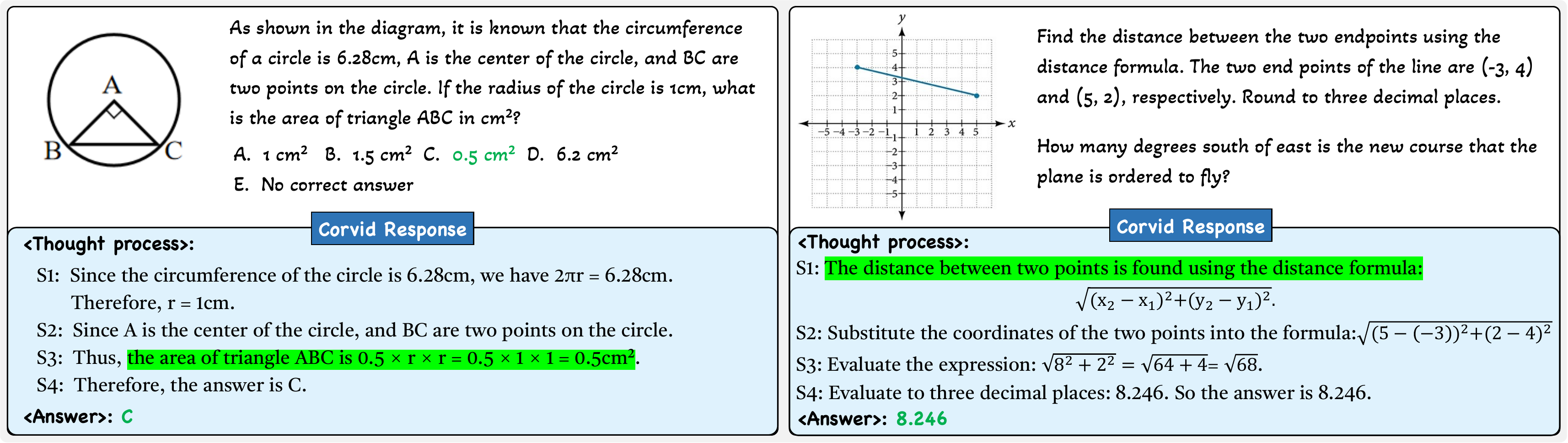}
% \vspace{-1mm}
\caption{The CoT responses generated by \Ours in mathematical reasoning.}
\label{fig:vis_math}
\end{subfigure}%
\vfil
% \vspace{1mm}
\begin{subfigure}{\linewidth}
\centering
\includegraphics[width=0.99\linewidth]{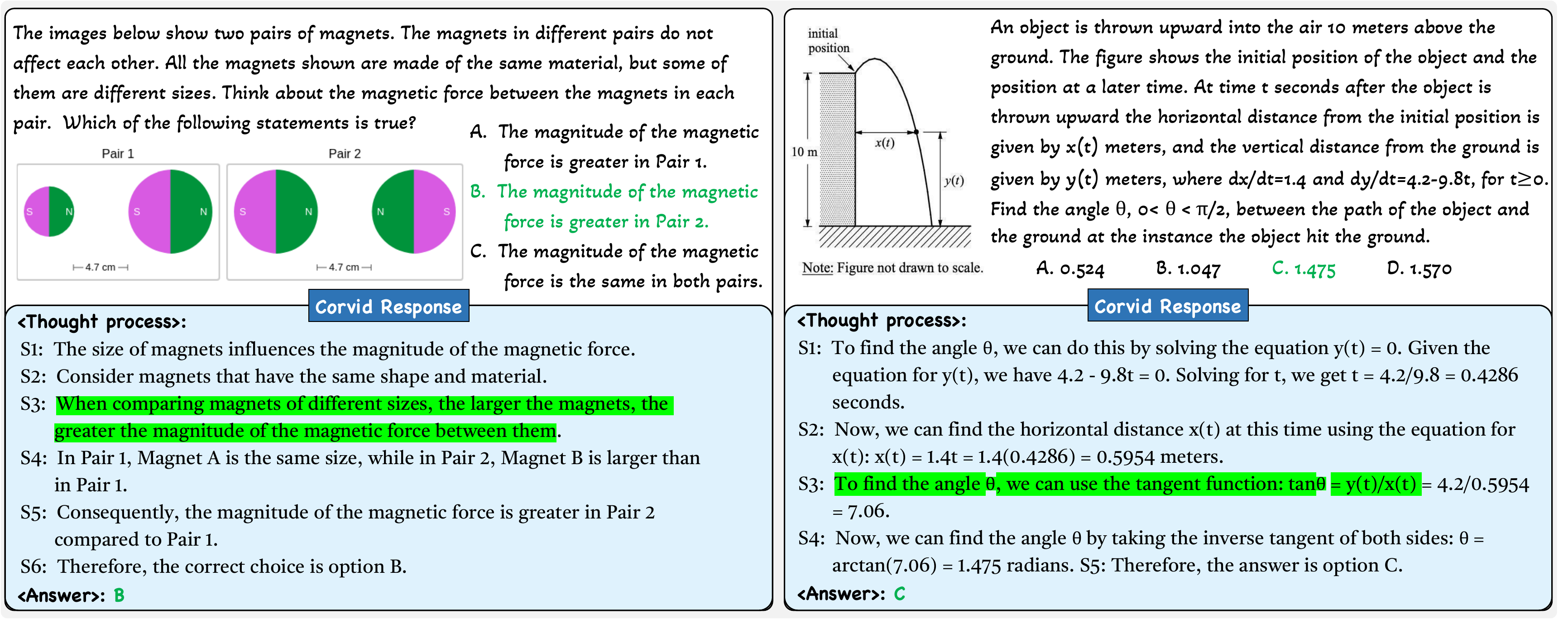}
% \vspace{-1mm}
\caption{The CoT responses generated by \Ours in science problem-solving.}
\label{fig:vis_sqs}
\end{subfigure}
\vfil
% \vspace{1mm}
\begin{subfigure}{\linewidth}
\centering
\includegraphics[width=0.99\linewidth]{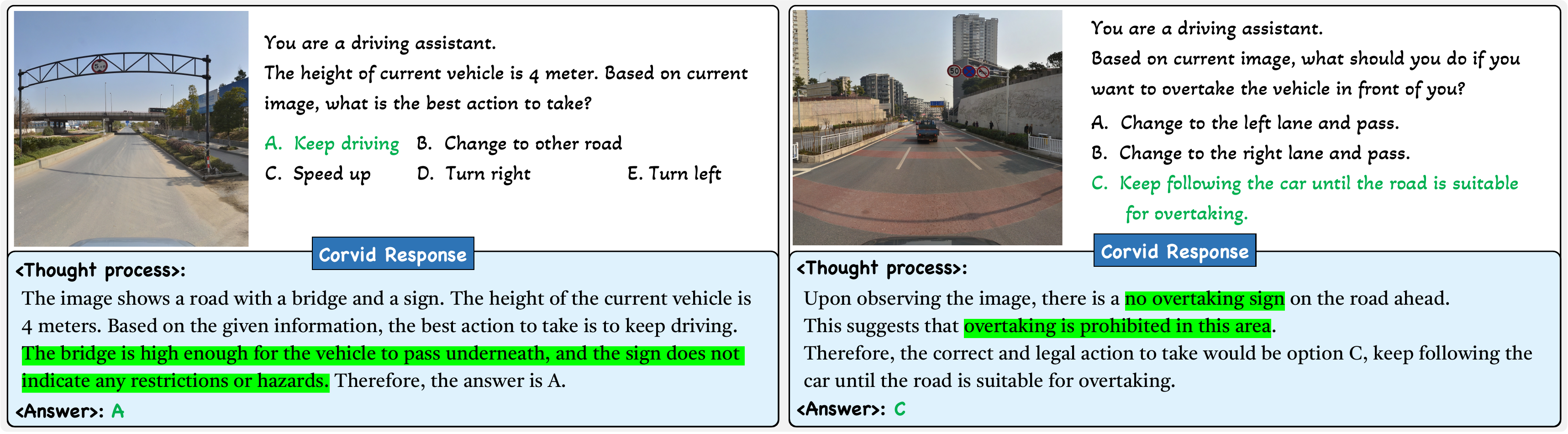}
% \vspace{-1mm}
\caption{The CoT responses generated by \Ours in visual comprehension.}
\label{fig:vis_vision}
\end{subfigure}
\vspace{-3mm}
\caption{
\textbf{Qualitative result of \Ours} in performing mathematical reasoning, science problem-solving, and visual comprehension. 
}
\vspace{-3mm}
\label{fig:app_vis}
\end{figure*}
% *********************************************************

\begin{figure*}[!t]
\centering 
\includegraphics[width=0.95\linewidth]{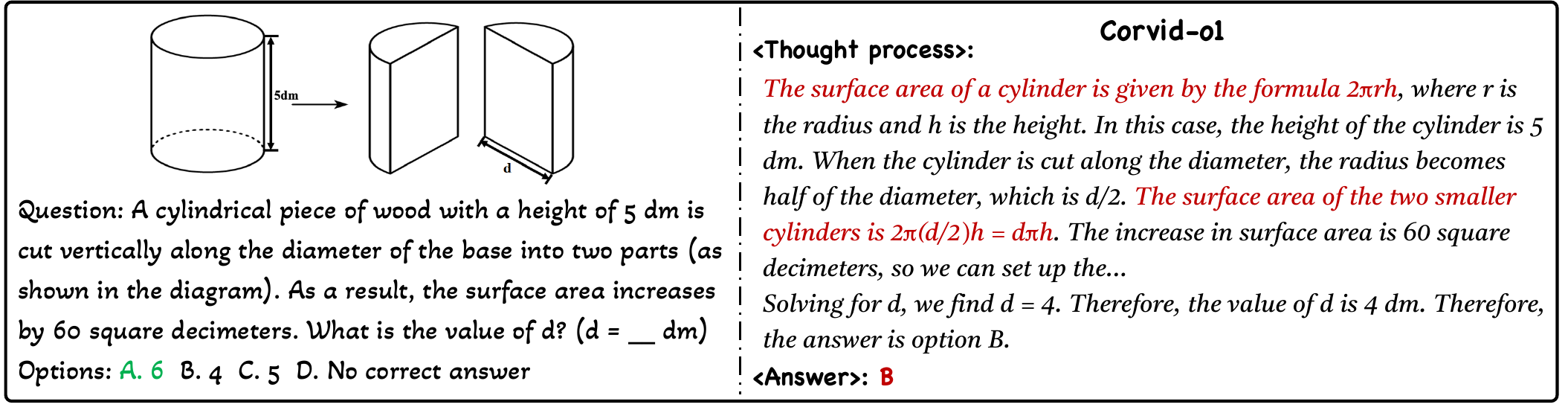}
\vspace{-3mm}
\caption{\textbf{A failure case} where \Ours-o1 fails to reason effectively due to insufficient domain-specific knowledge.}
\label{fig:failure_case_supp}
\vspace{-2mm}
\end{figure*}
\begin{figure*}[!t]
\centering 
\includegraphics[width=0.95\linewidth]{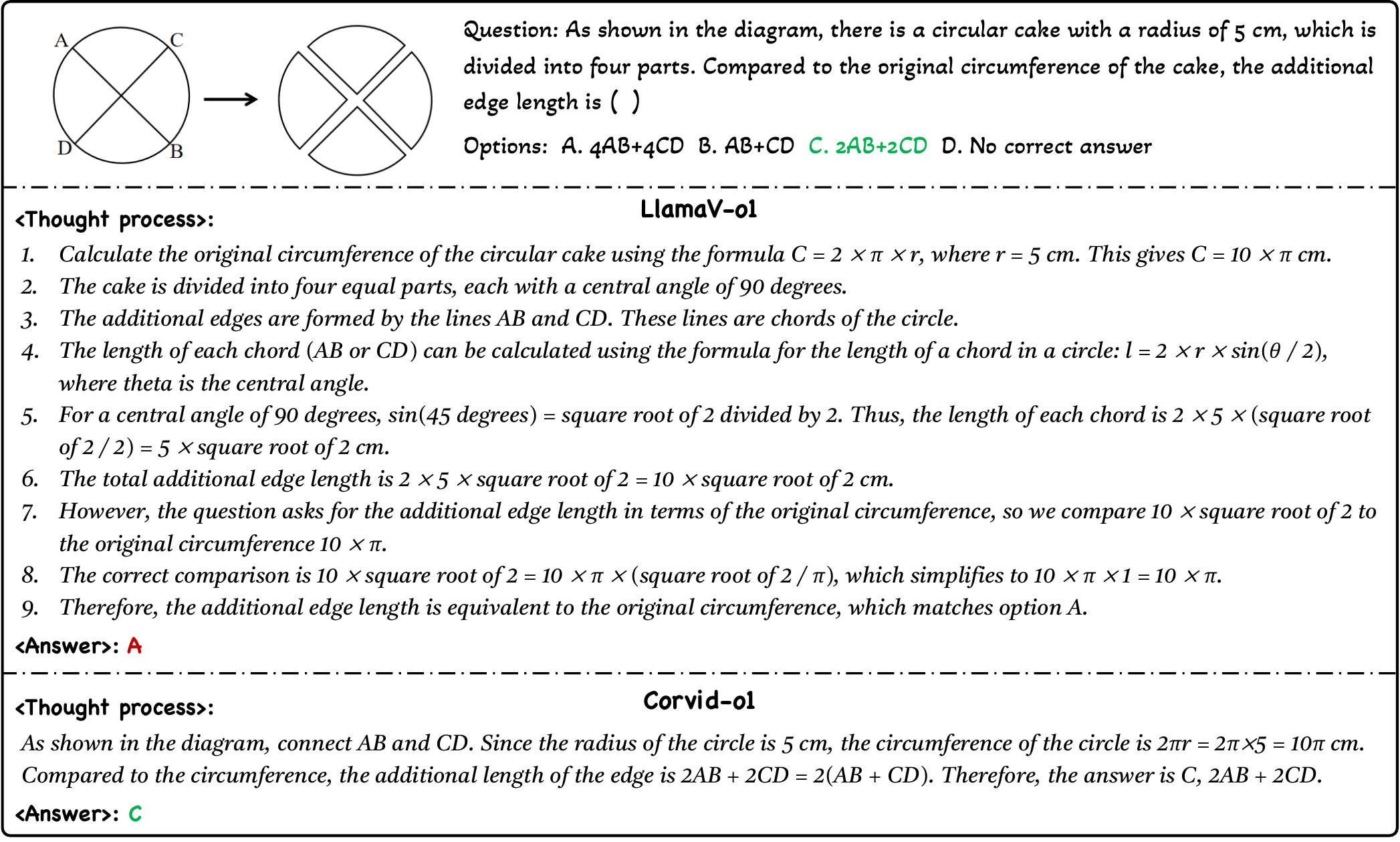}
\vspace{-3mm}
\caption{\textbf{Qualitative comparison} between LlamaV-o1 and \Ours-o1 in  mathematical reasoning.}
\label{fig:vis_comp_math}
\vspace{-2mm}
\end{figure*}

% **********************************************************

\subsection{Inference Efficiency}
\label{app:infer_efficiency}

To evaluate the efficiency and effectiveness of our inference-time scaling strategy, we compare the average inference time and accuracy of our Corvid, LLaVA-o1, and LlamaV-o1 on the MathVista benchmark comprising 1,000 test instances. 
LLaVA-o1 and LlamaV-o1 utilize stage-level beam search and sentence-level beam search, respectively. Following their optimal configurations, we set the beam size to 2 for LLaVA-o1 and 4 for LlamaV-o1. 
\cref{tab:infer_efficiency} shows that Corvid achieves significantly lower inference latency while maintaining higher accuracy. Specifically, Corvid-o1 reduces the inference time per instance to 11.4 seconds, yielding a 4.4$\times$ speedup over LLaVA-o1 and a 1.35$\times$ speedup over LlamaV-o1, while also surpassing both models in accuracy. 
This improvement is primarily attributed to Corvid-o1's more streamlined and simplified intermediate reasoning process during inference, as its inference time grows linearly with the number of generated tokens. 

% *********************************************************
\begin{table}[!hbt]
\centering
\footnotesize 
\setlength{\tabcolsep}{1.35mm}{
\begin{tabularx}{\linewidth}{lcccc} 
\toprule 
MLLMs 
&LLaVA-o1
&LlamaV-o1
&\Ours-o1$^\dagger$
&\Ours-o1-8B
\\
\midrule 
\rowcolor{gray!20}
Time (second)
&50.6
&15.4
&11.4
&11.3
\\ 
Accuracy
&56.1
&54.4
&61.5
&72.0
\\ 
\bottomrule
\end{tabularx}
}
\vspace{-3mm}
\caption{\textbf{The average inference time per instance} on MathVista, evaluated using a single NVIDIA A800 (80G) GPU.}
\label{tab:infer_efficiency}
\vspace{-3mm}
\end{table}
% *********************************************************

% **********************************************************
\subsection{Qualitative Results}
\label{app:exp_vis}

In \cref{fig:app_vis}, we provide an intuitive understanding of \Ours's CoT reasoning capabilities. 
As illustrated, when performing science and math reasoning, as well as visual comprehension, \Ours-o1 consistently generates faithful and detailed thought processes before arriving at an answer, enhancing the reliability and interpretability of its answer and demonstrating exceptional CoT capabilities.

\subsection{Additional Failure Case}
In addition to the case shown in \cref{fig:app_case}, \cref{fig:failure_case_supp} presents a typical failure case in mathematical reasoning, where \Ours-o1 fails to arrive at the correct answer due to insufficient domain-specific knowledge.

\subsection{Qualitative Comparison}
\begin{figure*}[h]
\centering 
\includegraphics[width=0.95\linewidth]{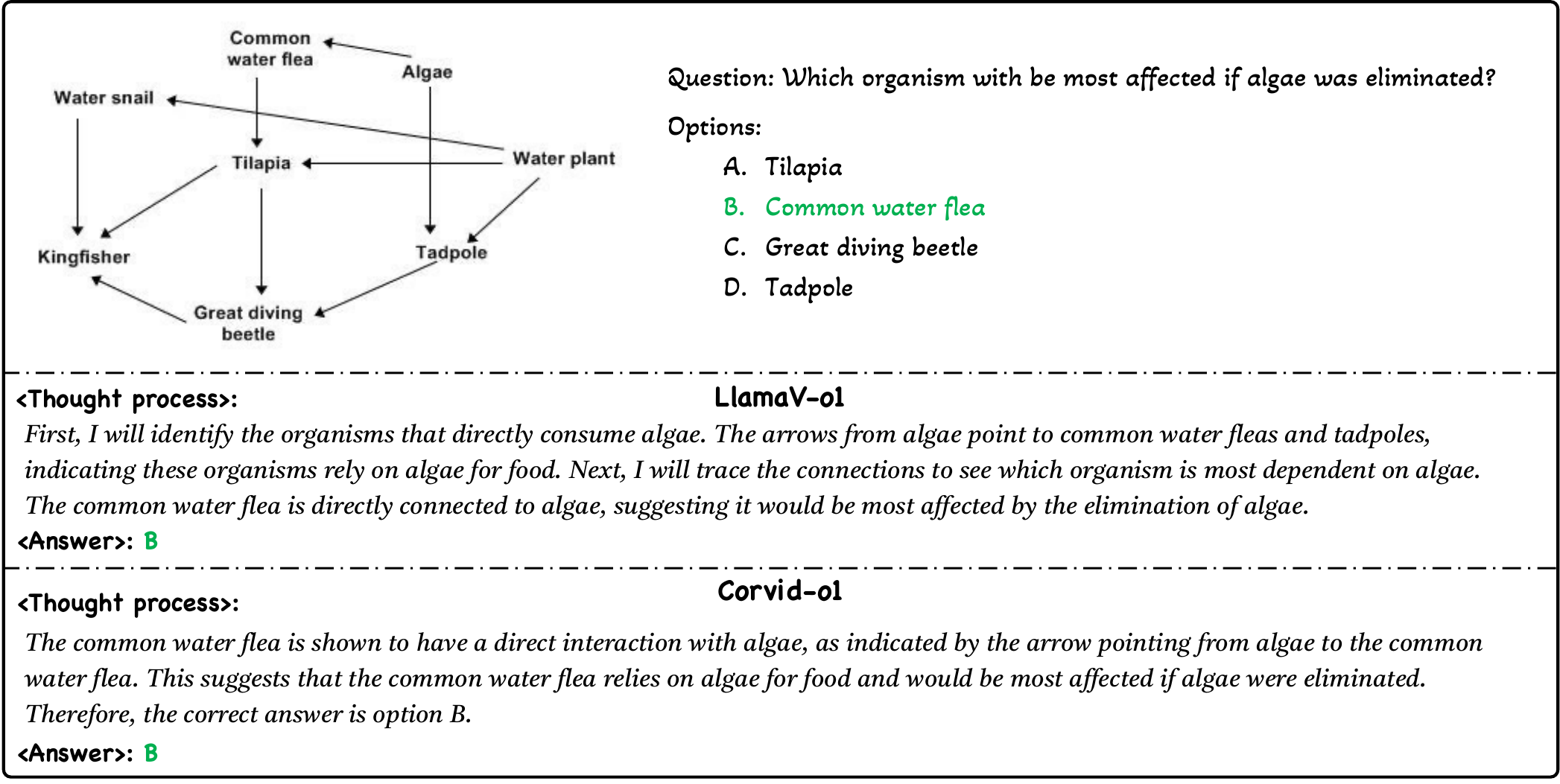}
\vspace{-3mm}
\caption{\textbf{Qualitative comparison} between LlamaV-o1 and \Ours-o1 in science problem-solving.}
\label{fig:vis_comp_science}
\vspace{-2mm}
\end{figure*}

\begin{figure*}[h]
\centering 
\includegraphics[width=0.95\linewidth]{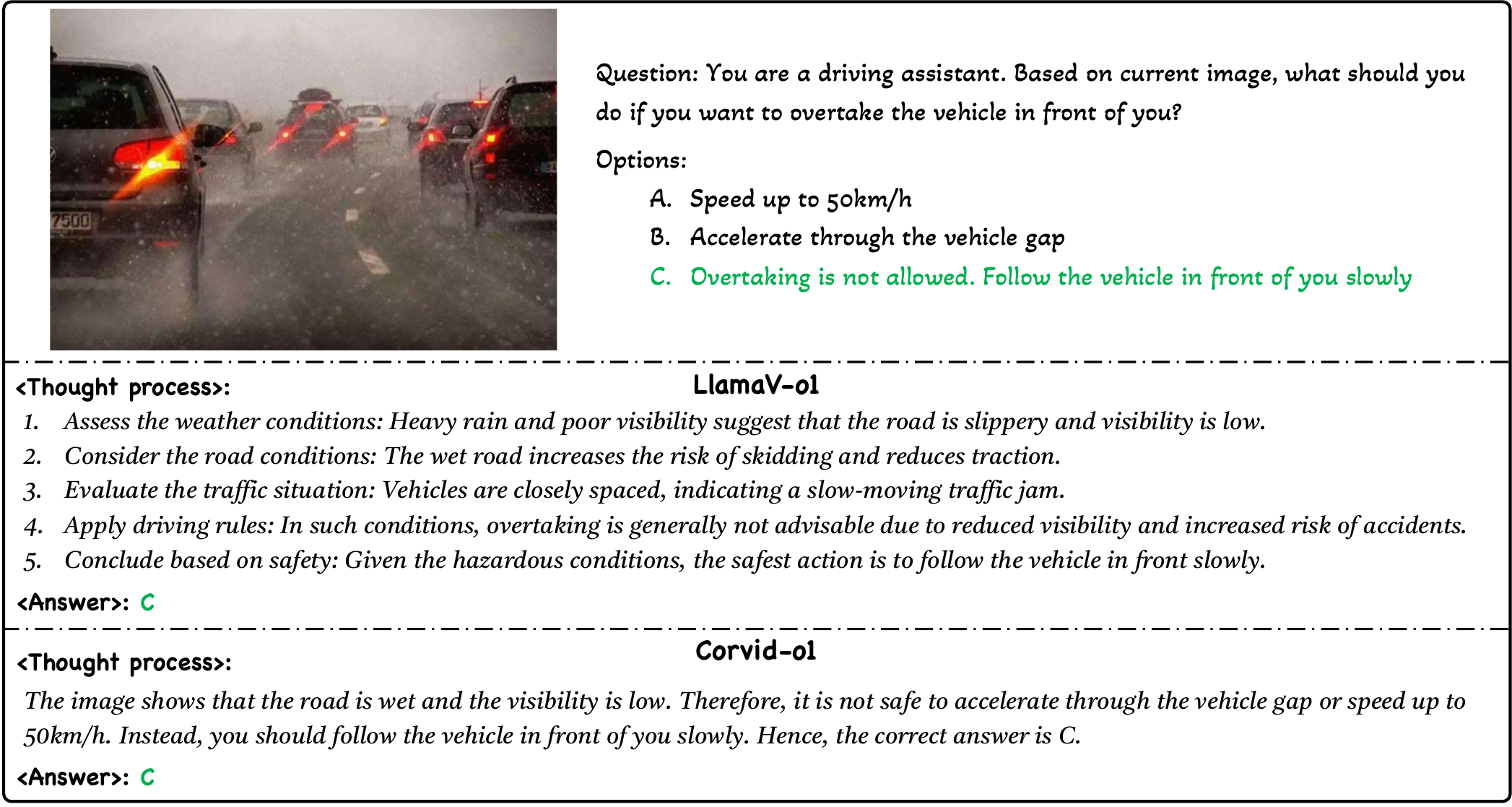}
\vspace{-3mm}
\caption{\textbf{Qualitative comparison} between LlamaV-o1 and \Ours-o1 in visual comprehension.}
\label{fig:vis_comp_und}
\vspace{-2mm}
\end{figure*}

\cref{fig:vis_comp_math,fig:vis_comp_science,fig:vis_comp_und} visualize several qualitative comparisons between LlamaV-o1 and \Ours-o1 across tasks.

% **********************************************************
% ********************************************************

\section{\mmcot}
\label{app:mcot_instruct}

In this section, we introduce MCoT-Instruct-287K, our high-quality multimodal CoT instruction-following dataset. Specifically, we first describe its sources and then elaborate the process of improving the quality of raw CoTs.

% ********************************************************
\subsection{Source of Raw Data}

As detailed in \cref{tab:app_mcot_instruct_source}, we collect data from seven manually created reasoning datasets and three AI-assisted generated reasoning datasets, totaling 292K raw instances spanning diverse reasoning types and domains, to construct MCoT-Instruct. 
Although all datasets provide initial rationales that serve as CoT responses, significant quality issues exist: AI-assisted generated CoTs may contain errors and duplications, while manually-created CoTs are usually brief and logically incoherent, rendering the raw data too noisy and unstandardized for effective CoT-enhancement training.

% ********************************************************
\subsection{Improving the Quality of Raw CoT}

% ********************************************************

% ************************************************************************** 
\begin{table}[h]
\centering
\footnotesize
\renewcommand{\arraystretch}{1.2}
\setlength{\tabcolsep}{1.2mm}{
% \resizebox{\linewidth}{!}{
\begin{tabular}{llr}
\toprule
Reasoning Type &Raw Dataset &Size 
\\ 
\midrule 
\rowcolor{gray!20}
\ding{172} General visual reasoning 
&GPT-VQA~\cite{zhao2023mllm} &26K  
\\ 
\ding{173} Knowledge-intensive visual reasoning 
&A-OKVQA~\cite{schwenk2022okvqa} &18K 
\\
\rowcolor{gray!20}
\ding{174} Visual Commonsense Reasoning
&VCR~\cite{zellers2019recognition} &84K 
\\

\multirow{3}{*}{\ding{175} Science Problem-Solving}
&M$^3$CoT~\cite{chen2024m} &9K 
\\ 
&SQA-IMG (train)~\cite{lu2022learn} &8K 
\\ 
% Scientific Comprehension
&ArxivQA~\cite{li2024multimodal} &54K 
\\ 
\rowcolor{gray!20}
&GeomVerse~\cite{kazemi2023geomverse} &9K 
\\ 
\rowcolor{gray!20}
\multirow{-2}{*}{\ding{176} Geometric Reasoning}
&R-CoT~\cite{deng2024r} &53K 
\\ 
% Geometric 
\ding{177} Numerical Reasoning
&GeoQA~\cite{chen2021geoqa} &7K
\\
\rowcolor{gray!20}
% Tabular-based Math Word Problems 
\ding{178} Mathematical reasoning 
&TabMWP~\cite{lu2023dynamic} &24K
\\ 
% MM-Math~\cite{sun2024advancing} &5K 
% \\ 
\bottomrule
\end{tabular}
}
\vspace{-3mm}
\caption{\textbf{Raw data of MCoT-Instruct.} Here, GPT-VQA, R-CoT, and ArxivQA are the three AI-assisted generated datasets.}
\label{tab:app_mcot_instruct_source}
\vspace{-3mm}
\end{table}

\begin{figure}[!t]
\centering 
\includegraphics[width=0.99\linewidth]{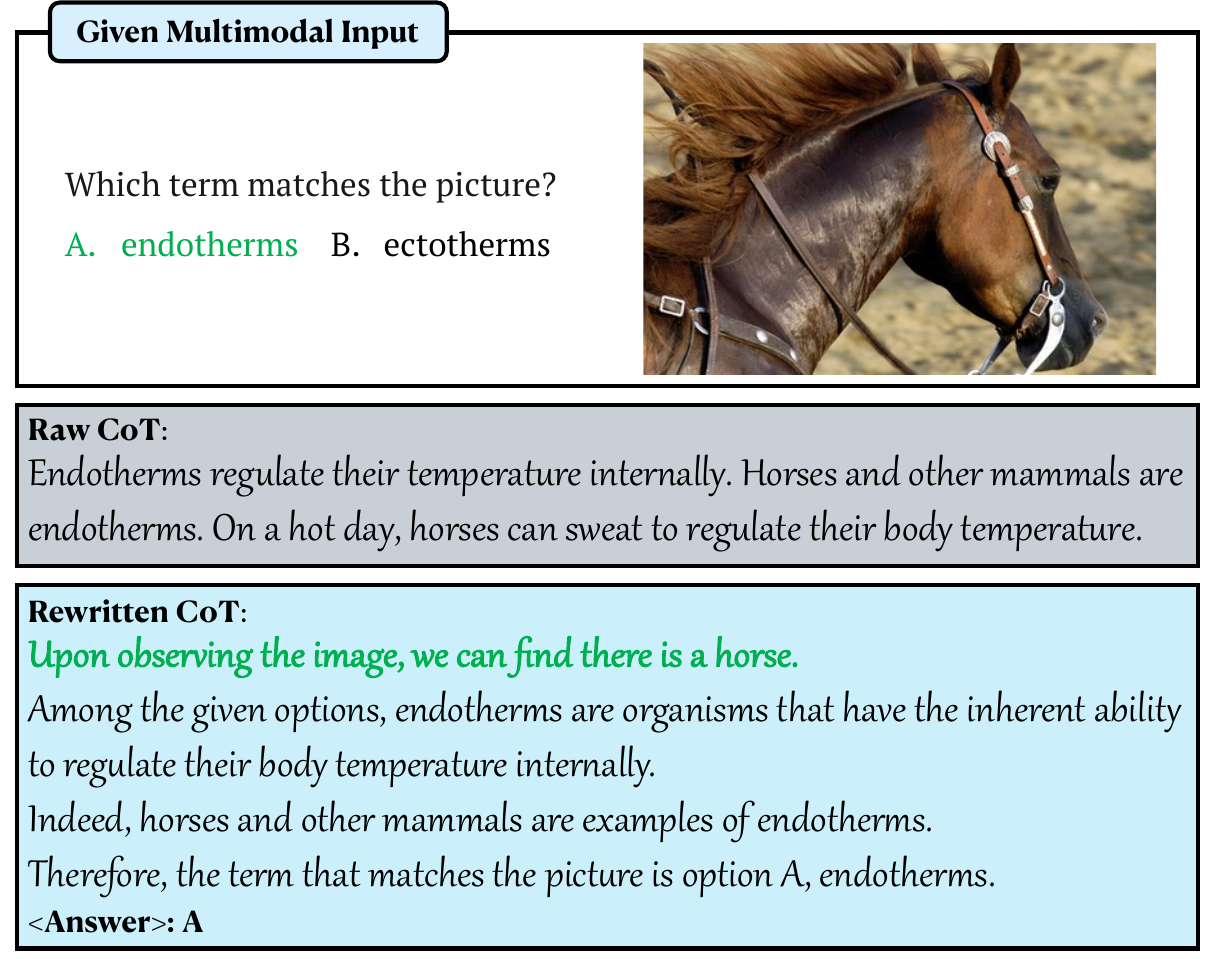}
\vspace{-3mm}
\caption{\textbf{Comparison between raw and rewritten CoTs.}}
\label{fig:mcot_expample}
\vspace{-3mm}
\end{figure}
% ********************************************************

To improve the quality of raw CoTs, we separately refine and standardize the aforementioned manually created and AI-generated reasoning datasets with GPT assistance through the following two steps: 
\begin{enumerate}[label={$\bullet$},topsep=2pt,itemsep=2pt] 
\item \textbf{CoT Rewriting.\,} 
As shown in \cref{fig:rat_rewrite}, we design a specialized prompt to instruct GPT-4o to refine these raw CoTs from manually-created datasets, enhancing their diversity and logical consistency. 
As demonstrated in \cref{fig:mcot_expample}, the rewritten CoTs remain faithful and consistent with the given context while becoming more detailed, logically coherent, and standardized. 
\item \textbf{Quality Verification and Data Filtering.\,} 
To guarantee the quality of all rewritten CoTs and those from AI-assisted generated datasets, we employ GPT to evaluate free-text CoTs across three dimensions: \textit{faithfulness}, \textit{relevance}, and \textit{completeness}. Inspired by the success of LLMs in automatic evaluation~\cite{chiang2023closer,liu2023gpteval}, we design a base prompt, as shown in \cref{fig:rat_eval}, to instruct GPT-4o to assign an overall score (0 - 1) to each rewritten CoT and its corresponding raw CoT. The CoT with the higher score is selected as the high-quality CoT. After that, we filter out instances with an overall score below 0.6. 
\end{enumerate}

With these steps, we ultimately obtain 287K instance with high-quality CoT responses that are consolidated into single-turn conversation instances of \mmcot. Notably, no testing or validation instances from any evaluation benchmark were included in this process. We used only the training split of ScienceQA for data curation, and \Ours's in-domain performance was evaluated exclusively on its respective test set.

% colback=green!5,colframe=green!35!black,
\begin{figure*}[!t]
\centering
\footnotesize
\begin{tcolorbox}[width=0.98\linewidth,arc=8pt,left=8pt,right=8pt,top=5pt,bottom=5pt,before skip=0.15cm,after skip=0.15cm,colback=bk-tcolorbox,]
% \begin{tcolorbox}[width=\linewidth,arc=2pt,left=2pt,right=2pt,top=2pt,bottom=2pt,before skip=0.15cm,after skip=0.15cm,colback=bk-tcolorbox,]
\hlp{\textbf{System message}} \\
You are an AI assistant that can do text rewritten. \\
\rule[4pt]{\linewidth}{0.05em}

\hlp{\textbf{Prompt}} \\
I want you to act as a Chain-of-Thought (CoT) Rewriter. Given a question with several options and its CoT response (i.e., the intermediate reasoning steps or rationales that lead to the correct answer), your objective is to rewrite the given CoT into a more standardized version.

\vspace{6pt}
\textbf{The rewritten CoT must follow the following rules:}
\begin{enumerate}[label={\arabic*)},topsep=1pt,itemsep=0pt,leftmargin=5mm]
\item Keep the logic of reasoning-then-answering to ensure that the reasoning can be performed step by step. 
\item Be faithful enough to ensure that the reasoning can accurately lead to the correct answer.
\item Be clear and concise, without factual errors or repeated content, and no key intermediate reasoning steps are omitted.
\item Do not mention or refer to the given CoT in your responses directly.
\end{enumerate}

\vspace{6pt}
\textbf{You can rewrite the given CoT using the following methods:}
\begin{enumerate}[label={\arabic*.},topsep=1pt,itemsep=0pt,leftmargin=5mm]
\item Improve existing reasoning steps or rationales to make the CoT more coherent and smooth.
\item Add more intermediate reasoning steps or rationales to make the CoT more specific and detailed.
\end{enumerate}

\vspace{6pt}
{\it Please make sure you have read and understood these instructions carefully.}

\vspace{6pt}
Following is an exemplar: 

\myboxedtext[azure]{
\textbf{\#\#\# Given question \& options:}\\
Which person is in charge?\\
A. person3 [0.1231, 0.1601, 0.4296, 0.9848] is the leader \\ 
B. person2 [0.5440, 0.1151, 0.7528, 0.9990] is in charge \\
C. person3 [0.1231, 0.1601, 0.4296, 0.9848] seems to be the one in charge \\ 
D. person3 [0.1231, 0.1601, 0.4296, 0.9848] has the higher rank \\
\textbf{\#\#\# Given CoT:}\\
person3 [0.1231, 0.1601, 0.4296, 0.9848] is walking ahead of the other two and speaking while person1 [0.3602, 0.0979, 0.5477, 0.9689] looks to be more of a worker. \\
\textbf{\#\#\# Rewritten CoT:}\\
The person who is in charge can be determined by observing their behavior and position. In this case, person3 [0.1231, 0.1601, 0.4296, 0.9848] is walking ahead of the other two individuals and is also the one who is speaking, indicating that they have a leadership role. On the other hand, person1 [0.3602, 0.0979, 0.5477, 0.9689] appears to be more of a worker. Based on these observations, it can be concluded that person3 seems to be the one in charge. \\ 
Hence, the correct answer is B.
}

\vspace{6pt}
Here is the example to be rewritten:

\textbf{\#\#\# Given Question \& Options:} 

$\{\}$

\textbf{\#\#\# Given CoT:} 

$\{\}$

% \vspace{6pt}
\textbf{\#\#\# Rewritten CoT:}

\vspace{6pt}
Now you can start to rewrite the given CoT.
\end{tcolorbox}
% \vspace{-4mm}
\caption{\textbf{Prompt template of CoT rewriting.} Here, we use one instance from the VCR dataset~\cite{zellers2019recognition} as the exemplar. 
For other source datasets, replace the given exemplar with data-specific instances. 
}
\label{fig:rat_rewrite}
\end{figure*}

% colback=green!5,colframe=green!35!black,
\begin{figure*}[!t]
\centering
\footnotesize 
\begin{tcolorbox}[width=0.95\linewidth,arc=8pt,left=8pt,right=8pt,top=5pt,bottom=5pt,before skip=0.15cm,after skip=0.15cm,colback=bk-tcolorbox,]

\hlp{\textbf{System message}} 
\vspace{2pt}

You are a helpful AI assistant that can evaluate the quality of free-text chain-of-thought (CoT) responses generated by a multimodal large language model (MLLM). \\
\rule[4pt]{\linewidth}{0.05em}

\hlp{\textbf{Prompt}}
\vspace{2pt}

You will be provided with the input context to the MLLM (i.e., an image description, a question, and several options for the question), along with the corresponding CoT response generated by the MLLM. 
Your task is to evaluate the free-text CoT responses and give a final overall score (0 - 1) based on the following three perspectives: 
\begin{enumerate}[label={\color{violet}{\ding{114}}},topsep=3pt,itemsep=-1pt,leftmargin=4.5mm]
\item \listnumber{\textbf{Faithfulness} (0 - 1):} it refers to how accurately the CoT response reflect the actual reasoning process of the MLLM. 
A faithful CoT response is one that genuinely represents the factors and logic the MLLM used to arrive at its answer. For example, if the MLLM generates an answer based on certain key points in the given context, a faithful CoT response would accurately describe how it picked those points and how they led to the answer. The focus of faithfulness is on the transparency and truthfulness of the explanation. 
\item \listnumber{\textbf{Relevance} (0 - 1):} it measures how the CoT response aligns with and supports the answer generated by the MLLM. A consistent CoT response should logically justify the answer, demonstrating a clear and direct connection between the CoT response and the inferred answer. That is, a consistent CoT response should not only be aligned with the answer but also provide sufficient and convincing reasons for why the answer is valid. 
\item \listnumber{\textbf{Completeness} (0 - 1):} it evaluates whether the CoT response provided by the MLLM encompasses all essential information and reasoning necessary to understand the MLLM's answer reasoning process. A complete CoT response should cover all critical aspects and steps of the MLLM's reasoning without omitting key details.
\end{enumerate}

\vspace{6pt}
{\textbf{Evaluation Steps:}} 
\begin{enumerate}[label={\arabic*.},topsep=2pt,itemsep=0pt,leftmargin=5mm]
\item Understand and analyze the provided image description, question, and options. 
\item Read the MLLM's response and systematically assess the CoT response from the three perspectives of Faithfulness, Relevance, and Completeness. 
\item Assign a final overall score (0 - 1) by averaging Faithfulness, Relevance, and Completeness.
\end{enumerate}

\vspace{6pt}
\textit{Please make sure you read and understand these instructions carefully.}
% Please keep this document open while reviewing, and refer to it as needed.

\vspace{6pt}
% \textbf{\#\#\# Example:}
The sample to be scored:\\
\textbf{\#\#\# Image Description:} \\ $\{\}$ \\
\textbf{\#\#\# Question \& Options:} \\ $\{\}$ \\
% \textbf{\#\#\# Options:} \\ $\{\}$ \\
\textbf{\#\#\# CoT Response:} \\ $\{\}$

\vspace{6pt}
\textbf{Evaluation Form:} \\
Answer by starting with ``Scoring:'' and then give the explanation of the score by ``Explanation:''

- Overall: 
\end{tcolorbox}
% \vspace{-4mm}
\caption{\textbf{Prompt template for CoT quality evaluation.} 
% (Score-Explain Auto-CoT Format)
}
\label{fig:rat_eval}
\end{figure*}

% ********************************************************

% ********************************************************
\section{Benchmark Details}

% ***************************************************************************
\begin{table}[!t]
\centering
\footnotesize 
\setlength{\tabcolsep}{1.4mm}{
\begin{tabular}{lccr} 
\toprule 
{Benchmarks}
&{Task Format}
&{Metric}
&{\#Sample}
% &\textbf{w/ CoT}
\\
\midrule 
% \multicolumn{5}{c}{$\blacktriangledown$~{\textit{problem solving \& multimodal reasoning}}}
% \\ 
MMStar~\cite{chen2024we}
&multi-choice &Accuracy &1,500 
\\ 
MMMU~\cite{yue2023mmmu}
&multi-choice &Accuracy &900 
\\ 
SQA-IMG~\cite{lu2022learn}
&multi-choice &Accuracy &2,017 
\\
AI2D~\cite{kembhavi2016diagram}
&multi-choice &Accuracy &3,088 
\\ 
% M$^3$CoT
% &multi-choice &Test &2,318 &\cmark
% \\ 
WeMath~\cite{qiao2024we}
&multi-choice &Accuracy &1,740 
\\ 
MathVista~\cite{lu2023mathvista}
&multi-choice\&free-text &Accuracy &1,000 
\\ 
MathVerse~\cite{zhang2024mathverse}
&multi-choice\&free-text &Accuracy &3,940 
\\ 
MathVision~\cite{wang2024measuring}
&multi-choice\&free-text &Accuracy &3,040 
\\ 
DynaMath~\cite{zou2024dynamath}
&multi-choice\&free-text &Accuracy &5,010 
\\ 

\midrule 
% \rowcolor{gray!20}
% \multicolumn{5}{c}{$\blacktriangledown$~{\textit{Comprehensive}}}
% \\ 
SEED-IMG~\cite{li2023seed}
&multi-choice &Accuracy &14,232 
\\
MMT-Val~\cite{ying2024mmt} 
&multi-choice &Accuracy &31,325 
\\
RWQA~\cite{xai2024grok}
&multi-choice &Accuracy &1901 
\\ 
BLINK~\cite{fu2024blink}
&multi-choice &Accuracy &1,901 
\\ 
\midrule
MMB~\cite{liu2023mmbench} 
&multi-choice &Accuracy &6,666 
\\ 
MMVet~\cite{yu2023mm} 
&free-text &GPT Score &218 
\\ 
Hallusion~\cite{guan2024hallusionbench}
&multi-choice &Accuracy &254 
\\ 
\bottomrule
\end{tabular}
}
\vspace{-3mm}
\caption{\textbf{Summary of evaluation benchmarks.}}
\label{tab:app_eval_bench}
\vspace{-3mm}
\end{table}

\cref{tab:app_eval_bench} presents all evaluation benchmarks used in this work. 
The task formats of MathVista, MathVerse, MathVision, and DynaMath encompass both multiple-choice question answering and free-text generation, while MMVet formats tasks as free-text generation. All other benchmarks are limited to multiple-choice question answering. Each benchmark adopts accuracy as its primary metric, except for MMVet, which utilizes a GPT-based score. 
Notably, SQA-IMG includes human-annotated CoTs, serving as references for assessing the quality of model's CoT responses in \cref{tab:result_cot_quality}.

\end{document}